\documentclass[10pt,twocolumn,letterpaper]{article}

\usepackage[accsupp]{axessibility}  
\usepackage{iccv}
\usepackage{times}
\usepackage{epsfig}

\usepackage{graphicx}
\usepackage{amsmath}
\usepackage{amssymb}
\usepackage{booktabs}

\usepackage{multirow}
\usepackage{multicol}
\usepackage{dsfont}
\usepackage{colortbl}
\usepackage{lipsum}
\usepackage{xspace}
\usepackage{makecell}
\usepackage{color}
\usepackage{enumitem}
\usepackage{amsthm}
\usepackage{wasysym}
\usepackage{wrapfig}
\usepackage{algorithm}
\usepackage{algorithmicx}
\setitemize{noitemsep,topsep=0pt,parsep=0pt,partopsep=0pt}

\newcommand{\nncat}[0]{6\xspace}
\newcommand{\nnpretrain}[0]{9947\xspace}

\newcommand{\vpara}[1]{\vspace{0.07in}\noindent\textbf{#1}\xspace}

\definecolor{mygray}{gray}{.9}
\definecolor{mypink}{rgb}{.99,.91,.95}
\definecolor{mygreen}{rgb}{.52,.73,.30}
\definecolor{myblue}{rgb}{.39,.58,.93}
\definecolor{mycyan}{cmyk}{.3,0,0,0}
\definecolor{mylakeblue}{rgb}{.0,.749,1.}
\definecolor{mypurple}{rgb}{.729,.333,.827}
\definecolor{myclassicblue2}{rgb}{.117,.565,1.}
\definecolor{myclassicblue}{rgb}{1.0,.3176,.3176}
\definecolor{myorange}{rgb}{1.0,.647,0}
\definecolor{kanki}{rgb}{.941,.902,.549}
\definecolor{myyellow}{rgb}{1.,.8745,.1745}
\definecolor{mygray2}{gray}{.5254}
\usepackage[pagebackref=true,breaklinks=true,letterpaper=true,colorlinks,bookmarks=false,citecolor=myblue]{hyperref}

\usepackage[capitalize]{cleveref}
\crefname{section}{Sec.}{Secs.}
\Crefname{section}{Section}{Sections}
\Crefname{table}{Table}{Tables}
\crefname{table}{Tab.}{Tabs.}

\usepackage[noend]{algpseudocode} 

\iccvfinalcopy 

\def\iccvPaperID{9013} 

\begin{document}

\title{Few-Shot Physically-Aware Articulated Mesh Generation\\via Hierarchical Deformation}


\author{
Xueyi Liu$^{1,5}$,
Bin Wang$^{2}$,
He Wang$^{3}$,
Li Yi$^{1,4,5}$
\\
$^{1}$Tsinghua University~~
$^{2}$Beijing Institute for General Artificial Intelligence~~
$^{3}$Peking University~~
\\
$^{4}$Shanghai Artificial Intelligence Laboratory~~
$^{5}$Shanghai Qi Zhi Institute
}

\maketitle
\thispagestyle{empty}



\begin{abstract}
    We study the problem of few-shot physically-aware articulated mesh generation. 
    By observing an articulated object dataset containing only a few examples, we wish to learn a model that can generate diverse meshes with high visual fidelity and physical validity. 
    Previous mesh generative models either have difficulties in depicting a diverse data space from only a few examples or fail to ensure physical validity of their samples. 
    Regarding the above challenges, we propose two key innovations, including 1) a hierarchical mesh deformation-based generative model based upon the divide-and-conquer philosophy to alleviate the few-shot challenge by borrowing transferrable deformation patterns from large scale rigid meshes and 2) a physics-aware deformation correction scheme to encourage physically plausible generations. 
    We conduct extensive experiments on \nncat articulated categories to demonstrate the superiority of our method in generating articulated meshes with better diversity, higher visual fidelity, and better physical validity over previous methods in the few-shot setting. 
    Further, we validate solid contributions of our two innovations in the ablation study. 
    Project page with code is available at \href{https://meowuu7.github.io/few-arti-obj-gen/}{meowuu7.github.io/few-arti-obj-gen}. 
\end{abstract}

\section{Introduction}

\begin{figure}[ht]
    \centering
    \includegraphics[width=0.45\textwidth]{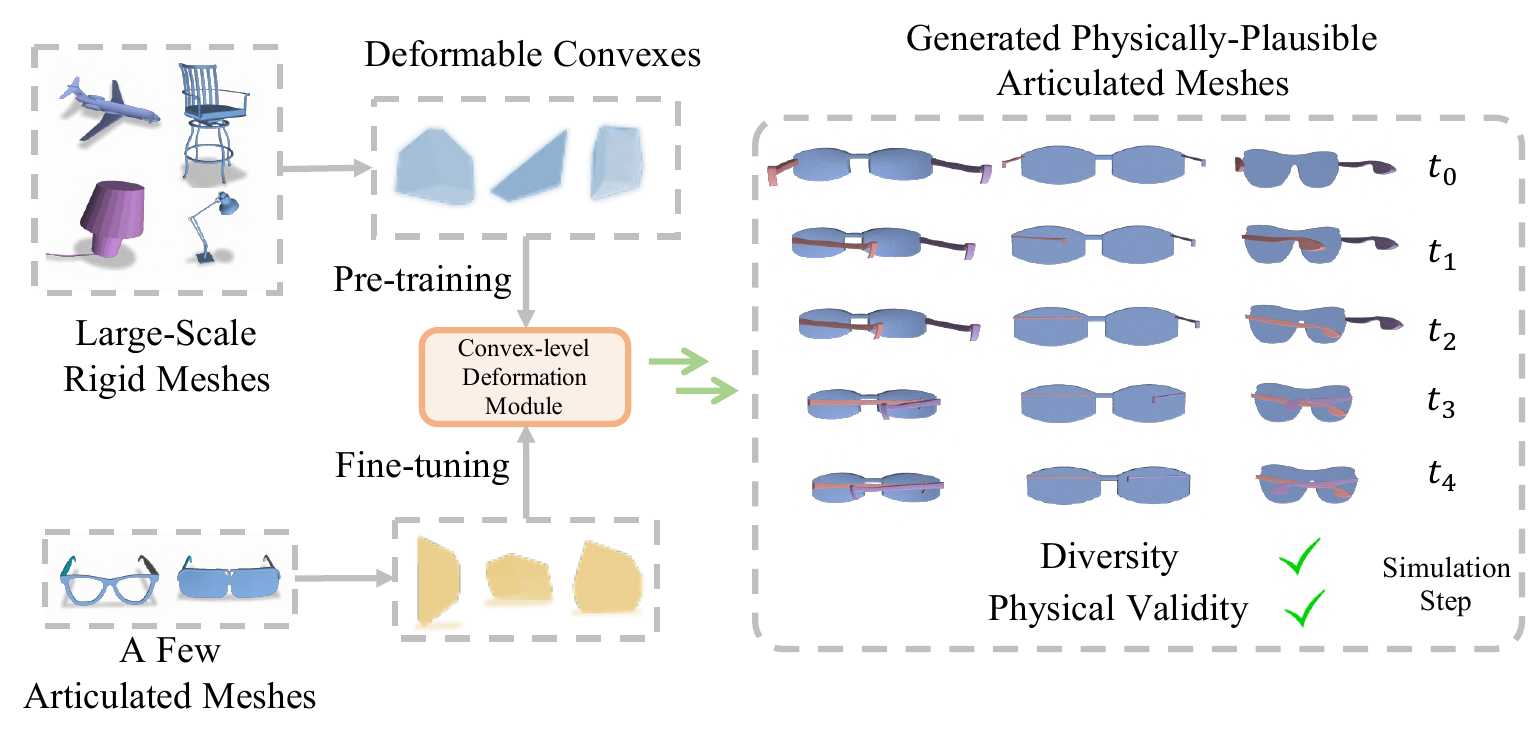}
    \caption{ \footnotesize
    \textbf{Overview.}
    We present a hierarchical mesh deformation-based generative model to solve the challenging yet important few-shot physically-aware articulated mesh generation problem. 
    It tackles the few-shot challenge by borrowing shared convex level deformation patterns from large-scale rigid meshes and incorporates a deformation correction scheme to further enhance the model's ability to generate physically realistic meshes. 
        }
        \vspace{-20pt}
    \label{fig_teaser}
\end{figure}

Generative models have aroused a wide spectrum of interests in recent years for their creativity and broad downstream application scenarios~\cite{rombach2022high,saharia2022photorealistic,singer2022make,lan2022dream,gao2022get3d,nash2020polygen}.
Specific to 3D generation, a variety of techniques such as denoising diffusion~\cite{luo2021diffusion,zhou2021pvd,chou2022diffusionsdf,yang2019pointflow} have also been discussed for a while. 
Among them,  mesh generation is indeed important since the mesh representation can support a wider range of downstream applications such as rendering and physical simulation compared to other representations such as point clouds. Existing works mainly focus on generating meshes for whole objects~\cite{gao2022get3d,nash2020polygen,chou2022diffusionsdf,li2022diffusion,saharia2022photorealistic} considering without modeling object functionalities at all. 
Besides, they mainly rely on reconstructing meshes from other kinds of representations such as implicit fields~\cite{gao2022get3d,chou2022diffusionsdf,li2022diffusion} instead of generating meshes directly. 
In this work, we go one step further and consider mesh generation for articulated objects that can support physically realistic articulations. This not only helps understand the object distribution in real-world assets, but also allows an intelligent agent to learn segmenting~\cite{li2020category,liu2023self}, tracking~\cite{weng2021captra}, reasoning~\cite{hong2022fixing} and manipulating~\cite{xu2021umpnet} articulated objects through a simulation environment.
We focus on the articulated mesh generative model that can generate object meshes with diverse geometry, high visual fidelity, and correct physics.

Training a generative model on publicly available articulated mesh datasets to depict a diverse physically-plausible data space not limited to training assets presents two main challenges to the methodology. 
First, existing articulated object datasets are usually very restricted in scale. For example, the PartNet-Mobility Dataset~\cite{xiang2020sapien} contains an average of 51 meshes per category. 
This naturally requires a few-shot generative model to learn from a very limited number of meshes. 
Adapting previous approaches immediately without carefully considering the few-shot nature would lead to models suffering from poor generative ability. Second, we need to pursue physically plausible generation to ensure the generated meshes are not only visually appealing but also functionally sound to support correction articulation functions,  \emph{i.e.,} attached parts without self-penetrations in the full articulation range.

Despite recent advancements in mesh generation community such as a wide variety of models proposed in existing works~\cite{gao2022get3d,zeng2022lion,chou2022diffusionsdf,shue20223d,li2022diffusion}, they are typically challenged by the following difficulties and always fail to solve our problem:
1) Lack of the ability to learn a wide data space not limited to training shapes in the few-shot setting. 
2) Difficulty in modeling crucial object-level shape constraints imposed by the functionality of articulated objects. Failure to consider these requirements would result in physically unrealistic samples~\cite{gao2022get3d,chou2022diffusionsdf,nash2020polygen}.
Modeling such physical constraints for articulated meshes is a non-trivial task, as it requires accounting for diverse penetration phenomena caused by different types of articulation motions.
To our best knowledge, we are the first that presents a valid framework to address such two difficulties for articulated mesh generation.

Our work designs a hierarchical mesh deformation-based generative model that tackles the aforementioned challenges using two key innovations:
(1) Hierarchical mesh deformation with transfer learning. We introduce an object-convex shape hierarchy and learn the hierarchical articulated mesh generative model.
The model is trained by first learning the deformation-based generative model at the leaf convex level and then synchronizing individual convex-level deformation spaces at the root level. 
We identify that different categories tend to share convex-level deformation patterns and 
leverage this insight to learn and transfer  
rich deformation prior from large-scale rigid datasets to expand the model's generative capacity. 
(2) Physics-aware deformation correction. To address self-penetrations of deformed articulated meshes during mesh articulation, we further introduce a deformation correction scheme. 
It is composed of an auxiliary loss penalizing self-penetrations during mesh articulation and a collision response-based shape optimization strategy. 
By integrating this scheme into the hierarchical mesh deformation model, we successfully guide the model to generate more physically realistic deformations, resulting in physically correct articulated meshes finally.

We conduct extensive experiments on \nncat categories from the PartNet-Mobility dataset~\cite{xiang2020sapien} for evaluation. 
As demonstrated by both the quantitative and qualitative results, we can consistently outperform all baseline methods regarding the fidelity, diversity and physical plausibility of generated meshes, \emph{e.g.,} an average of 10.4\% higher coverage ratio, 43.7\% lower minimum matching distance score, and 26.5\% lower collision score.
Ablation studies further validate the value of our design in deformation pattern transfer learning, the hierarchical mesh generation approach, and the effectiveness as well as the versatility of our physics-aware correction scheme. 

\textbf{Our key contributions} are as follows:
\textbf{(1)} We present the first solution, to our best knowledge, for the challenging yet important few-shot physically-aware articulated mesh generation problem with two effective and non-trivial technical innovations. 
\textbf{(2)} We propose a hierarchical mesh deformation-based generative model based upon the divide-and-conquer philosophy.
This design allows us to learn a diverse data space by borrowing shared deformation patterns from large-scale rigid object datasets.
\textbf{(3)} We propose a physics-aware deformation correction scheme to encourage the hierarchical generative model to produce physically realistic deformations, resulting in improved physical validity of the generated samples. 
This scheme can also be effectively integrated into other deformation-based mesh generative models, 
thereby enhancing the physical validity of their samples as well. 


\section{Related Works}


\vpara{Mesh generative models.}
There have been vast and long efforts in devising 3D mesh generative models~\cite{nash2020polygen,gao2022get3d,zeng2022lion,chou2022diffusionsdf,liu2021deepmetahandles,wen2019pixel2mesh}. 
Their techniques can be mainly categorized into three genres:
1) direct surface generation~\cite{nash2020polygen}, 2) deformation-based mesh generation~\cite{liu2021deepmetahandles,wen2019pixel2mesh}, and 3) hybrid representation-based generation~\cite{gao2022get3d,zeng2022lion}. 
Though methods of the first type exhibit obvious merits such as synthesizing high-quality n-gon meshes, they always suffer from limited generative ability and cannot scale for complex objects. 
In contrast, deformation-style mesh generation models deform source shapes for new samples which naturally spares the efforts for mesh structure generation, while restricted by poor  flexibility.
The third strategy separates the mesh surface structure generation
problem from the content generation, which offers them with powerful generative ability.  
However, the quality of their samples is coupled with the power of their surface reconstruction techniques~\cite{peng2021shape,shen2021deep}. 
In this work, we leverage mesh deformation as our generation technique for articulated mesh synthesis, taking advantages of its ability to produce high-quality samples. 
Instead of deforming whole objects or parts directly, we design a hierarchical deformation strategy to enhance the deformation flexibility and to enrich the data space by borrowing deformation patterns shared across categories from large scale rigid meshes. 

\vpara{Few-shot generation.} 
Along with the flourishing image generative models emerged in recent years, the few-shot image generation has been widely explored as well~\cite{hong2022deltagan,gu2021lofgan,hong2020matchinggan,hong2020f2gan,antoniou2017data}. 
It wishes to create more data given only a few examples from a novel category that is both diverse in content and semantically consistent with the target category. At the high level, their basic philosophy is to design proper approaches such that the model can benefit from large base datasets for generation, like local fusion~\cite{gu2021lofgan}, latent variables matching~\cite{hong2020matchinggan,bartunov2018few}. adversarial delta matching~\cite{hong2022deltagan}. 
%
In this work, we leverage transfer learning to adapt shape patterns from large-scale rigid datasets to target articulated categories. 
Devising methods in this way requires us to find correct intermediates on which shape patterns are cross-category transferrable. Instead of directly using whole objects or articulated parts, we choose convexes as such intermediates. 
Transferring knowledge at this level presents further difficulties in fusing them together in a geometrically consistent way and in synthesizing physically realistic meshes while mainting visual diversity at the same time. 

\vpara{Physics-aware machine learning.}
Our work is also related to physics-aware machine learning~\cite{ren2023diffmimic,hong2022fixing,Hu2022physicalinteraction,mezghanni2022physical,shu20203d,mezghanni2021physically,li2017grass}, and mostly relevant to physically-aware generative models. 
To ensure physical validity of generated shapes, typical solutions  choose either offline simulations for training data filtering~\cite{shu20203d} or online simulations leveraging the development of differentiable simulators~\cite{de2018end,hu2019difftaichi,hu2019chainqueen} or by designing online simulation layers~\cite{mezghanni2022physical}. 
Generating physically-plausible articulated objects presents new challenges considering self-penetrations during mesh articulation that are more complex than stability issues caused by gravity for rigid objects. 
Our method integrates the physical supervision and a shape optimization strategy. 
The optimization transforms part shapes to resolve self-penetration issues. 
High-dimensional and complex shape deformations are involved in the process, different from linear fixing operations considered in~\cite{hong2022fixing}.  

\section{Method}

\begin{figure*}[ht]
    \centering
      \includegraphics[width=1.0\textwidth]{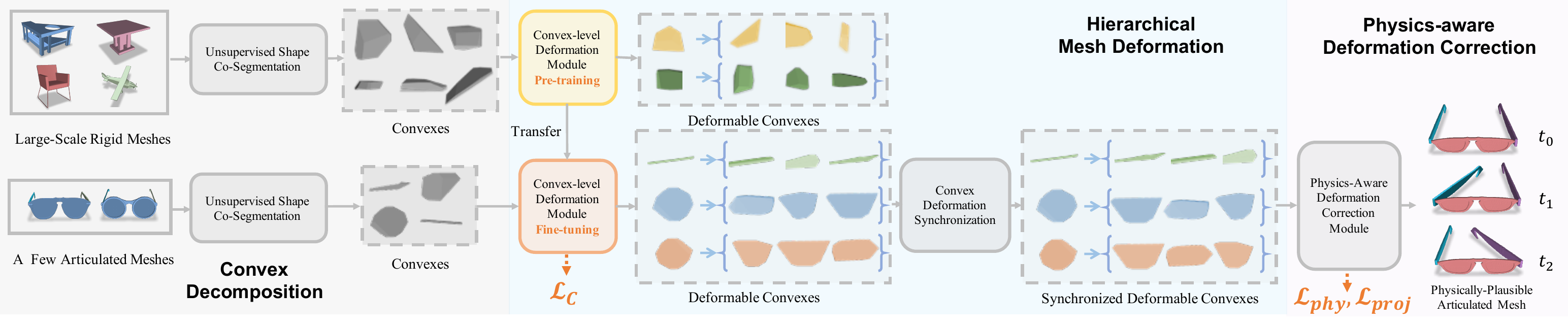}
    \caption{ \footnotesize
    \textbf{Framework overview for our few-shot physically-aware  articulated mesh generation.}
    In this figure, \textcolor{myyellow}{yellow} blocks represent modules with learnable weights optimized during pre-training. \textcolor{myorange}{Orange} blocks contain weights optimized during fine-tuning. \textcolor{mygray2}{Gray} blocks contain no learnable weights. Convexes of the same color are of the same type. Our framework consists of a hierarchical mesh deformation scheme that learns and transfers diverse shared deformation patterns from large-scale rigid datasets at the convex level. We also propose a convex deformation synchronization strategy to combine individual convex-level deformation spaces into the object-level space. Furthermore, we introduce a physics-aware deformation correction strategy to address self-penetrations in synthesized articulated meshes.
    }
    \label{fig_overall_pipeline}
\end{figure*}

The problem we are targeting is the few-shot physically-aware articulated mesh generation. Given a set of articulated meshes from the category of interest, we would like to learn a conditional generative model which could deform an articulated mesh from the same category into a wide variety of shapes. 
This conditional generation setup allows generating a large number of physically-plausible articulated meshes from a few examples while avoiding the need to generate the mesh structure.
However, it leaves several challenges to address including how to accurately represent complex shape deformation spaces from a few examples and how to ensure that the generated meshes support physically-realistic articulations.

Regarding the first challenge, our idea is to learn the deformation space via borrowing knowledge from other object categories. This seemingly simple idea is not trivial to realize though since we need to figure out what knowledge is transferrable. We present a hierarchical mesh deformation strategy to allow deformation prior to transfer at a local convex segment level while still maintaining the deformation consistency at the global shape level. Regarding the second challenge, we introduce a physics-aware deformation correction scheme to avoid unwanted artifacts such as self-penetrations during mesh articulation.

In the following, we will provide a pipeline overview in Section~\ref{sec:overview}. Then we will explain our hierarchical mesh deformation strategy and our physics-aware deformation correction scheme in Section~\ref{sec:hier_deform} and Section~\ref{sec:phys_corr} respectively.

\subsection{Overview}
\label{sec:overview}
Given a small set of articulated meshes $\mathcal{A}$ from a certain category of interest with the same number of parts and joints, sharing a known kinematic chain, our method wishes to learn a conditional generative model 
depicting a diverse and plausible articulated shape space.
We adopt the divide-and-conquer philosophy and develop a hierarchical deformation scheme with transfer learning to tackle the difficulty of few-shot generation. 
Instead of learning at the whole shape level, we structure each articulated shape into an object-convex hierarchy and solve the generation problem via two steps. 
We first learn a generative model depicting a diverse shape space at the lowest convex level by borrowing common shape patterns from large rigid mesh datasets, denoted as $\mathcal{B}$. Convexes, with small cross-category distribution gap, serve as good intermediates for transferring common shape prior. After that, we devise a synchronization strategy that composes convex deformations consistently to form valid object shapes. 
Besides, a physics-aware correction scheme is developed to avoid physically-unnatural phenomena such as self-penetrations. Our overall pipeline is shown in Figure~\ref{fig_overall_pipeline}. 

Specifically, we first decompose the conditional mesh $a$ into approximately convex segments $\mathcal{C}_a$, forming an object-convex hierarchy. Then, on the leaf level, we learn to deform each convex $c\in\mathcal{C}_a$ through a convex-level conditional generative model $g_{\mathcal{C}}(\mathbf{z}_c|c)$ where $\mathbf{z}_c$ is the noise parameter corresponding to convex $c$. Finally, on the root level, we synchronize the convex deformations by replacing the convex-dependent noise parameter $\mathbf{z}_c$ with $S_c\mathbf{z}$, where $S_c$ is a linear transformation and $\mathbf{z}$ is a synchronized noise parameter shared among all convexes. This aligns the noise space of different convexes to form a coherent deformation for the whole mesh. The above hierarchical deformation strategy can be mathematically represented as $g(\mathbf{z}|a)=\{g_{\mathcal{C}}(S_c\mathbf{z}|c)|c\in \mathcal{C}_a\}$. 

During the training time, we first leverage existing unsupervised shape segmentation tools BSP-Net~\cite{chen2020bsp} to decompose meshes in both $\mathcal{A}$ and $\mathcal{B}$ into approximately convex segments $\mathcal{C}_{\mathcal{A}}$ and $\mathcal{C}_{\mathcal{B}}$. Since BSP-Net decomposes shapes consistently for each category, we can naturally identify corresponding convexes within $\mathcal{C}_{\mathcal{A}}$ or $\mathcal{C}_{\mathcal{B}}$. 
We can then pre-train a convex-level conditional generative model $g_{\mathcal{C}}(\mathbf{z}_c|c)$ on $\mathcal{C}_{\mathcal{B}}$ modeling how corresponding convexes could deform into each other among the large-scale rigid meshes. We then fine-tune the pre-trained model $g_{\mathcal{C}}(\mathbf{z}_c|c)$ on convexes in $\mathcal{C}_{\mathcal{A}}$ and at the same time estimate the noise synchronization transformation $S_c$. 
We further exploit an additional physics-aware deformation correction scheme to improve the physical validity of generated articulated shapes. 
It consists of 1) an auxiliary loss penalizing self-penetrations to provide physical supervision and 2) a collision response-based shape optimization strategy to encourage the model to generate physically realistic meshes. 
The auxiliary loss is incorporated into the training pipeline. While  the shape optimization scheme functions at both the training time and the test time.

\subsection{Hierarchical Mesh Deformation}
\label{sec:hier_deform}
Given an input articulated mesh $a$ and its corresponding approximately convex segments $\mathcal{C}_a$, our hierarchical mesh deformation model $g(\mathbf{z}|a)=\{g_{\mathcal{C}}(S_c\mathbf{z}|c)|c\in C_a\}$ consists of a convex-level conditional generative model $g_\mathcal{C}$ as well as a series of synchronization transformations $\{S_c\}$. 
We propose to learn the model at the lowest convex level $g_C$ following a transfer learning paradigm so that convex-level shape patterns can easily transfer across different categories. Given a reference mesh $a$, the hierarchical deformation first synthesizes new convex shapes via $g_C$. A deformation synchronization strategy is developed to handle the resulting deformation inconsistency issue across different convexes to form a valid object shape. 

\vpara{Convex-level conditional generative model.} We propose to leverage mesh deformation to characterize the convex-level conditional generative model. For a convex mesh segment $c$ containing $N_c$ vertices, the conditional generative model $g_\mathcal{C}(\mathbf{z}_c|c)$ should be able to produce diverse and realistic vertex-level deformation offset $\mathbf{d}_c\in \mathbb{R}^{N_c\times 3}$ when we sample different noise parameters $\mathbf{z}_c$.

The convex deformation $\mathbf{d}_c$ lies in a high-dimensional space which varies from convex to convex, prohibiting the knowledge transfer across different convexes. We therefore reparametrize $\mathbf{d}_c$ using two tricks inspired by~\cite{yifan2020neural, liu2021deepmetahandles}: 1) using cages to control per-vertex deformation; 2) using dictionaries to record the common deformation modes.

In particular, for each convex $c$, we use a coarse triangle mesh (a cage) $t_c$ enclosing $c$ to control the deformation of convex $c$~\cite{yifan2020neural}. The cage $t_c$ usually contains much less vertices compared with the convex $c$ so that its distribution is easier to be modeled. The deformation $\mathbf{d}_c$ of the convex $c$ can be easily computed as a linear transformation of the deformation $\mathbf{d}_{t_c}$ of cage $t_c$: $\mathbf{d}_c = \Phi_c \mathbf{d}_{t_c}$. Here $\Phi_c$ is an interpolation matrix based upon the generalized barycentric coordinates of convex $c$ with respect to cage $t_c$. We deform a template mesh based upon the shape of each convex to form the cages which we defer the details to supp.

To further reduce the deformation parametrization,  we represent the cage deformation $\mathbf{d}_{t_c}$ as a linear combination of $K$ deformation bases $B_c=[\mathbf{b}_c^{1}\;...\;\mathbf{b}_c^{K}]$ as $\mathbf{d}_{t_c}=B_c\mathbf{z}_c$, where $\mathbf{z}_c$ is a $K$-dimensional deformation coefficient. Here each deformation basis $\mathbf{b}_c$ represents a common deformation pattern and all the bases span the deformation space of cage $t_c$ and therefore convex $c$. Representing deformation spaces via deformation bases can effectively reduce the dimension of shape space compared to other alternatively such as utilizing latent vectors. 

\vpara{A few-shot deformation learning paradigm.} 
Given the above deformation reparametrizations, learning the convex-level conditional generative model for each convex $c$ boils down to learning the deformation bases $B_c$ as well as the distribution of deformation coefficient $\mathbf{z}_c$. 
For deformation bases, we employ a neural network $\psi_\theta(\cdot)$ to predict from convex shapes. 
It takes a given convex $c$ as input and outputs its deformation bases, \emph{i.e.,} $B_c = \psi_\theta(c)$. 
We then optimize the deformation coefficient $\mathbf{z}_c^{\hat{c}}$ for each convex $\hat{c}$ in correspondence to $c$ from the current available dataset. The distribution of $\mathbf{z}_c$ is then a mixture of Gaussian fit by the resulting deformation coefficients $\{ \mathbf{z}_c^{\hat{c}} \}$. 
We further leverage a transfer learning approach that transfers deformation priors learned in large-scale rigid dataset to target datasets at  the convex level based on the observation that the convex-level deformations usually show similar patterns across categories, \emph{e.g.,} a slab gets thicker or a strip gets enlongated. 
Therefore we can learn a diverse deformation space from a few examples. 

We pre-train $g_\mathcal{C}(\mathbf{z}_c\vert c)$ on the large-scale rigid mesh dataset $\mathcal{B}$ and fine-tune it on each target articulated dataset $\mathcal{A}$. 
In particular, at the pre-training time, given a set of rigid meshes $\mathcal{B}$ from large-scale online repositories as well as the corresponding convexes $\mathcal{C}_{\mathcal{B}}$, we first identify pairs of convexes in correspondence $\{(c, \hat{c})|c, \hat{c}\in \mathcal{C}_{\mathcal{B}}\}$ from the same-category shapes, \emph{e.g.,} the noses of two different airplanes. These correspondences come as a result of some off-the-shelf unsupervised co-segmentation algorithms~\cite{chen2020bsp}. 
We then optimize $g_\mathcal{C}(\mathbf{z}_c\vert c)$ by alternatively optimizing the deformation coefficient set $\{ \mathbf{z}_c^{\hat{c}}\vert c,\hat{c}\in \mathcal{C}_\mathcal{B} \}$, and the neural network $\psi_\theta(\cdot)$. 
To optimize $\{\mathbf{z}_c^{\hat{c}}\}$, we fix $\{ B_c \}$ and optimize each $\mathbf{z}_c^{\hat{c}}$ by minimizing the Chamfer Distance (CD), also denoted as $d_{\text{CD}}(\cdot, \cdot)$, between the deformed convex $c$ and the target $\hat{c}$. 
Then we optimize $\{ B_c \}$
by fixing deformation coefficients $\{ \mathbf{z}_c^{\hat{c}} \}$ and minimizing average CD between deformed $c$ and the target $\hat{c}$ for each pair $(c, \hat{c})$, which leads to the convex deformation loss $\mathcal{L}_C$ at the training time: 
\begin{equation}
    \mathcal{L}_C = \frac{1}{\vert \mathcal{C}_{\mathcal{B}} \vert} \sum_{c,\hat{c}\in \mathcal{C}_{\mathcal{B}}} d_{\text{CD}}(g_\mathcal{C}(\mathbf{z}_c\vert c, \mathbf{z}_c = \mathbf{z}_c^{\hat{c}}), \hat{c}).
\end{equation}
After the above alternative optimization, the distribution of  $\mathbf{z}_c$  for each convex $c$ is modeled by a mixture of Gaussian distribution fit to the final coefficient set $\{ z_c^{\hat{c}} \vert \hat{c}  \in \mathcal{C}_{\mathcal{B}} \}$. 
At the fine-tuning time, $g_\mathcal{C}(\mathbf{z}\vert c)$ is further optimized via the same procedure by the convex correspondence set  $\mathcal{C}_\mathcal{A}$ of the target dataset $\mathcal{A}$. 


\begin{figure}[ht]
    \centering
    \includegraphics[width=0.45\textwidth]{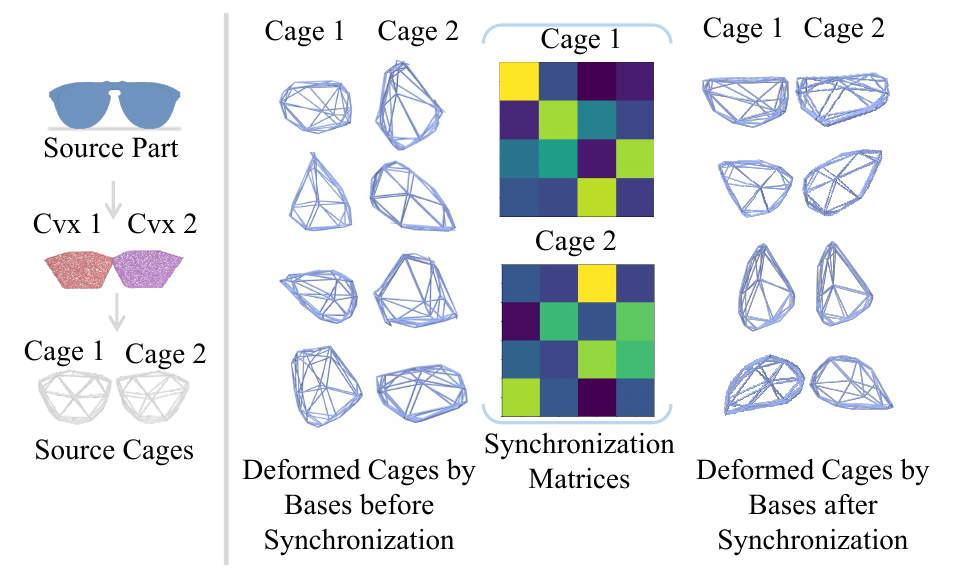}
    \caption{ \footnotesize
    \textbf{Synchronization Process.} The \emph{left part} illustrates the decomposed convexes and source cages of the input eyeglass frame. The \emph{right part} visualize synchronization matrices (a $4\times 4$ matrix here for each cage), cages deformed by bases before synchronization (\emph{left} two columns), and cages deformed by synchronized bases (\emph{right} two columns). 
        }
        \vspace{-16pt}
    \label{fig_sync_process}
\end{figure}

\vpara{Convex deformation synchronization.} 
After learning the conditional generative model for each convex $c$, the next step is to compose all the deformation spaces for the whole mesh $a$. Since for each convex $c$, $g_{\mathcal{C}}(\mathbf{z}_c|c)$ exploits a separate set of deformation bases $B_c$, the noise parameter $\mathbf{z}_c$ varies its meaning from convex to convex. As a result, if we draw independent noise parameters for each convex, the outcoming deformations could easily contradict with each other, failing the whole mesh-level deformation. 
To tackle this issue, we synchronize different deformation bases $B_c$ with linear transformations $S_c$ so that a single noise parameter $\mathbf{z}$ can be shared across all convexes.


%
Formally speaking, %
given a set of articulated object meshes $\mathcal{A}$ from a certain category and an articulated mesh $a\in\mathcal{A}$, assuming the mesh is segmented into $M$ convexes $\{c_m\}_{m=1}^M$ and each convex is equipped with a deformation model $g_{\mathcal{C}}(\mathbf{z}_{c_m}|c_m)$, our goal is to replace $\mathbf{z}_{c_m}$ with $S_{c_m}\mathbf{z}$ so that sampling the shared noise parameter $\mathbf{z}$ results in a globally consistent mesh deformation. To compute the synchronization transformation $S_{c_m}$, we consider the deformation from $a$ to other articulated meshes $a^i\in\mathcal{A}$. In particular, for each $a^i$, we optimize for a set of deformation coefficients $\{\mathbf{y}_m^i\}$ so that each convex $c_m$ in mesh $a$ could deform into the corresponding convex $c_m^i$ in mesh $a^i$ following the deformation model $g_{\mathcal{C}}(\mathbf{z}_{c_m}|c_m,\mathbf{z}_{c_m}=\mathbf{y}_m^i)$. 
We can then estimate the synchronization transformations $\{S_{c_m}\}$ by solving the following optimization problem:
\begin{align}
\vspace{-4ex}
    \underset{\{S_{c_m}\}, \{\mathbf{z}^i\}}{\text{minimize}} \sum_{i=1}^{\vert\mathcal{A}\vert}\sum_{m=1}^M\Vert B_{c_m}S_{c_m}\mathbf{z}^i-B_{c_m}\mathbf{y}_m^i\Vert_2,
    \label{eq_sync}
\end{align} 
\noindent where $B_{c_m}$ is the deformation bases of convex $c_m$ and $\mathbf{z}^i$ is a global deformation coefficient from mesh $a$ to $a^i$ shared across all convexes. We solve the above optimization problem via alternatively optimizing the synchronization transformations $\{S_{c_m}\}$ and the global deformation coefficients $\{\mathbf{z}^i\}$: 
\begin{itemize}
    \item Fix $\{ S_{c_m} \}$, optimize each global deformation coefficient $\mathbf{z}^i$ from $a$ to $a^i$ via Algorithm~\ref{algo_sync_opt_global_coeffs}.
    It takes the convex deformation bases $\{ B_{c_m} \}$, current synchronization transformations $\{ S_{c_m} \}$, convex deformation coefficients $\{ \mathbf{y}_m^i \}$ as input, and outputs the optimized $\mathbf{z}^i$. 
    \item Fix $\{  \mathbf{z}^i \}$,  optimize each synchronization transformation $S_{c_m}$ for each convex $c_m$ via Algorithm~\ref{algo_sync_opt_sync_matrix}. 
    It takes the  convex deformation bases $\{ B_{c_m} \}$, current global deformation coefficients $\{ \mathbf{z}^i \}$, convex deformation coefficients $\{ \mathbf{y}_m^i \}$ as input, and outputs the optimized $S_{c_m}$. 
\end{itemize}


\begin{figure}[ht]
  \vspace{-4ex}
\includegraphics[width=0.45\textwidth]{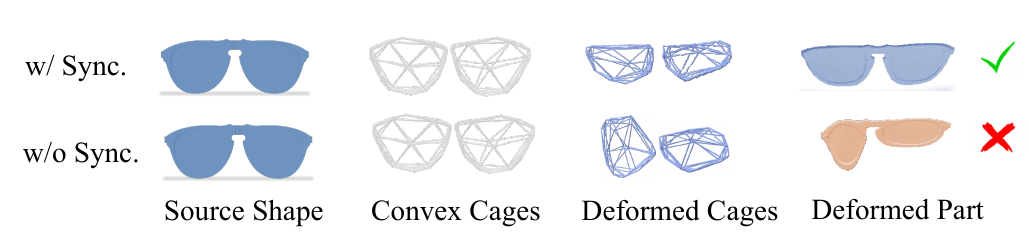}
  \caption{\footnotesize 
  \textbf{Synchronization's Effectiveness.} The synchronized deformation bases can consistently transform each convex for a valid part shape (upper row), while those before the synchronization fail (bottom row). 
  }
  \label{fig_sync_effect}
\end{figure}

\begin{algorithm}[H]
\small 
\caption{\textbf{Synchronization transformation matrices optimization}. 
}
\label{algo_sync_opt_sync_matrix}
\footnotesize
    \begin{algorithmic}[1]
        \Require
            Deformation bases for each convex $\{  B_{c_m}\}$. 
            Global deformation coefficients $\{ \mathbf{z}^i \}$ from $a$ to other articulated meshes $\{ a^i \}$. 
            Deformation coefficients $\{ \mathbf{y}_m^i \}$ from each convex $c_m$ to the corresponding convex of the articulated mesh $a^i$. 
        \Ensure
            Synchronization transformation matrix $S_{c_m}$ of the convex $c_m$.

        \State $\mathbf{Z} \leftarrow \text{Stack}(\{ \mathbf{z}^i \})$
        \State $\mathbf{Y}_m \leftarrow \text{Stack}( \{ \mathbf{y}_m^i \} )$
        \State $[\mathbf{U}, \mathbf{\Sigma}, \mathbf{V}^T] \leftarrow \text{SVD}(\mathbf{Z})$
        \State $[\mathbf{U}_m, \mathbf{\Sigma}_m, \mathbf{V}_m^T] \leftarrow \text{SVD}(\mathbf{Y}_m)$
        \State ${S}_{c_m} \leftarrow \mathbf{U}_m \mathbf{\Sigma}_m \mathbf{V}_m^T \mathbf{V} \mathbf{\Sigma}^+ \mathbf{U}^T$ \\
        \Return ${S}_{c_m}$
    \end{algorithmic}
\end{algorithm}

\begin{algorithm}[H]
\small 
\caption{\textbf{Global deformation coefficients optimization}. 
``lsq'' denotes the least square solver. 
}
\label{algo_sync_opt_global_coeffs}
\footnotesize
    \begin{algorithmic}[1]
        \Require
            Deformation bases for each convex $\{  B_{c_m}\}$. 
            Synchronization transformations $\{ S_{c_m} \}$. 
            Deformation coefficients $\mathbf{y}_m^i$ from each convex $c_m$ to the corresponding convex of the articulated mesh $a^i$. 
        \Ensure
            Global deformation coefficients $\mathbf{z}^i$ from $a$ to $a^i$. 

        \State $\mathcal{S}_{\mathbf{z}^i} \leftarrow \emptyset$
        \For{$m = 1$ to $M$}
            \State $\hat{\mathbf{z}}_m^i\leftarrow \text{lsq}({S}_{c_m}, \mathbf{z}_m^i)$
            \State $\mathcal{S}_{\mathbf{z}^i} \leftarrow \mathcal{S}_{\mathbf{z}^i}  \cup \{ \hat{\mathbf{z}}_m^i \}$
        \EndFor
        \State $\mathbf{z}^i = \text{Average}(\mathcal{S}_{\mathbf{z}^i})$ \\
        \Return $\mathbf{z}^i$
    \end{algorithmic}
\end{algorithm}

\noindent As an intuitive illustration of the synchronization process, we select the example of synchronizing the eyeglass frame's convex deformations (with 2 convexes and 4 deformation bases for each convex) and visualize the process (detailed to the cage level) in Figure~\ref{fig_sync_process}. The synchronized deformation bases can transform two convexes more consistently and symmetrically than those before synchronization (an example on deformed part shapes is shown in Figure~\ref{fig_sync_effect}).

After deformation synchronization, the distribution of the shared noise parameter $\mathbf{z}$ can be simply set as a mixture of Gaussian fit to the optimized global deformation coefficients $\{\mathbf{z}^i\}$.

\subsection{Physics-Aware Deformation Correction}
\label{sec:phys_corr}
Based upon the above design, our hierarchical deformation model can then synthesize a deformed mesh $a$ via the noise parameter $\mathbf{z}$ by taking a source articulated mesh as input. 
However, we may frequently observe physically unnatural self-penetrations when articulating $a$. 
To encourage the model to produce physically valid articulated meshes, we further propose a physics-aware deformation correction scheme,
which serves two purposes: 1) to optimize the weights of the generative model to produce deformations that are more physically realistic and 2) to optimize the shape $a$ to improve its physical validity. 
To accomplish this, we draw inspiration from previous works on stable rigid object generation~\cite{mezghanni2022physical,mezghanni2021physically} and develop an online articulation simulation process that places $a$ into $K$ different articulation states sequentially, denoted as $\text{Sim}(a) =  \{ a_k \}_{k=1}^K$. The articulation state sequence is designed by hand particularly for each category. 
We then utilize the physical supervision and a shape optimization strategy to guide the network to generate physically realistic articulated meshes. 
We will elaborate them in the following text. 




\vpara{Physical supervision.}
To provide physical supervision for articulated mesh generative models requires us to design stability signals to measure physically unstable phenomena, for which we mainly consider self-penetrations during mesh articulation. 
We therefore devise a metric named average penetration depth (APD) measuring the magnitude of each vertex penetrating through other parts during the articulation simulation process. 
It is also referred as  $\mathcal{L}_{phy}$ when we treat it as a loss. 
Formally, $\mathcal{L}_{phy} = \frac{1}{K}\sum_{k=1}^K \text{PeneD}(a_k)$, where $\text{PeneD}(a_k)$ measures self-penetrations in a single articulation state. 
We defer details of $\mathcal{L}_{phy}$ to the supp. 

One way to provide physical guidance for the network is to utilize  the physical stability signal  $\mathcal{L}_{phy}$  as an auxiliary loss to supervise the network training. 
However, physically unnatural phenomena of articulated meshes during mesh articulation are more diverse and complex than that of rigid objects caused by diverse part geometric appearance and wide articulation variations. 
Directly exploring physical supervision to guide the network optimization is not sufficient to regularize the network to produce physically realistic deformations. 
Therefore we further develop a collision response-based shape optimization strategy to improve the physical realism of the generated mesh $a$. 
Then we first optimize $a$ for several times to reduce self-penetrations and then use it to calculate $\mathcal{L}_{phy}$ for network optimization.


\vpara{Collision response-based shape optimization.} 
To realize the vision of resolving self-penetrations via optimizing shapes, we draw inspirations from collision response strategies and devise a heuristic penetration resolving strategy that projects penetrated vertices onto the surface of the mesh. 
To guide such projection, we devise an algorithm that calculates $\text{ProjD}(a)$ whose gradient over each penetrated vertex in $a$ informs how to project it to resolve penetrations. 
Then averaging the $\text{ProjD}(a_k)$ over each articulation state $\{ a_k\}$ yields the projection loss, \emph{i.e.,} $\mathcal{L}_{proj} = \frac{1}{K} \sum_{k=1}^K \text{ProjD}(\text{sim}_k(a))$. 
By iteratively using $\mathcal{L}_{proj}$ to update the global deformation coefficient $\mathbf{z}$ of the mesh, we can optimize the shape $a$ to mitigate self-penetrations. 
At the \textbf{training time}, $\mathcal{L}_{proj}$ is used to optimize deformation coefficients $\mathbf{z}$ for several iterations at first (\emph{i.e.,} 5 iterations), followed by leveraging $\mathcal{L}_{phy}$ calculated on the optimized shape to update network weights. 
\textbf{Test-Time Adaptation (TTA).} During the \textbf{test time}, only $\mathcal{L}_{proj}$ is iteratively applied to refine the result (\emph{i.e.,} for 10 iterations). 

\section{Experiments}

\begin{figure*}[ht]
    \centering
      \includegraphics[width=1.0\textwidth]{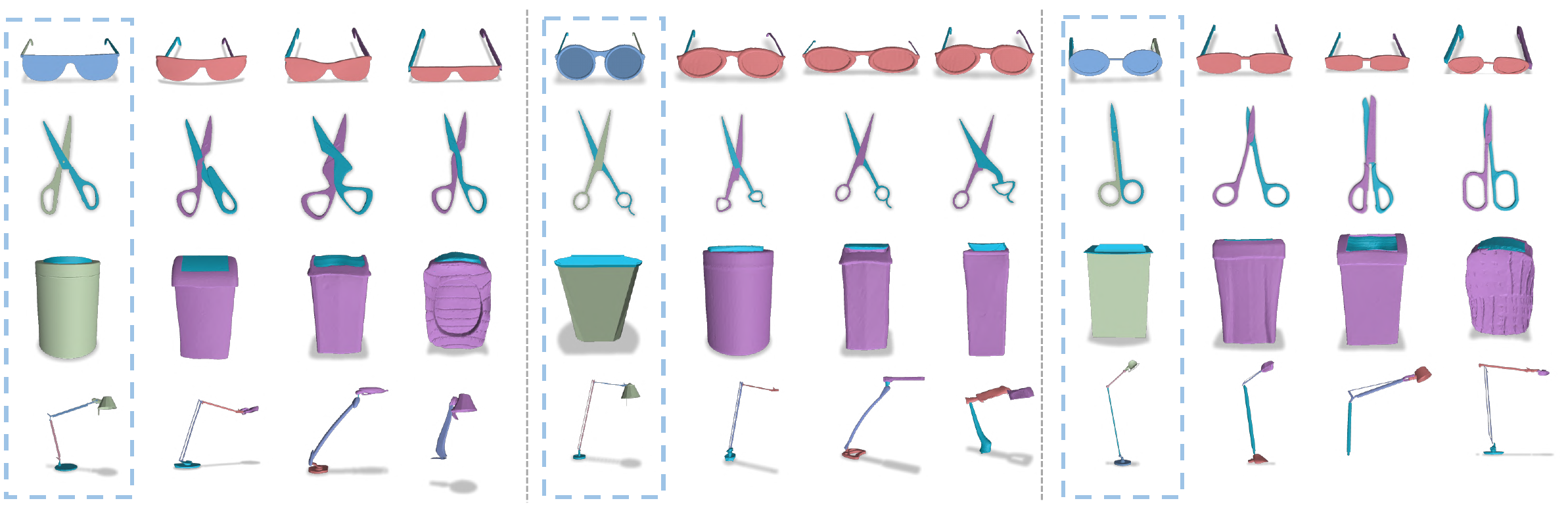}
    \caption{ \footnotesize
        \textbf{Qualitative evaluation on few-shot articulated mesh generation.}
        For every four shapes, the leftmost one (highlighted by \textcolor{myblue}{blue rectangles}) is the reference shape from the training dataset, while the remaining three are conditionally generated samples. Object categories from top to down are Eyeglasses, Scissors, TrashCan, and Lamp respectively. 
        }
        \vspace{-10pt}
    \label{fig_free_gen_vis}
\end{figure*}

\begin{figure}[ht]
    \centering
    \includegraphics[width=0.48\textwidth]{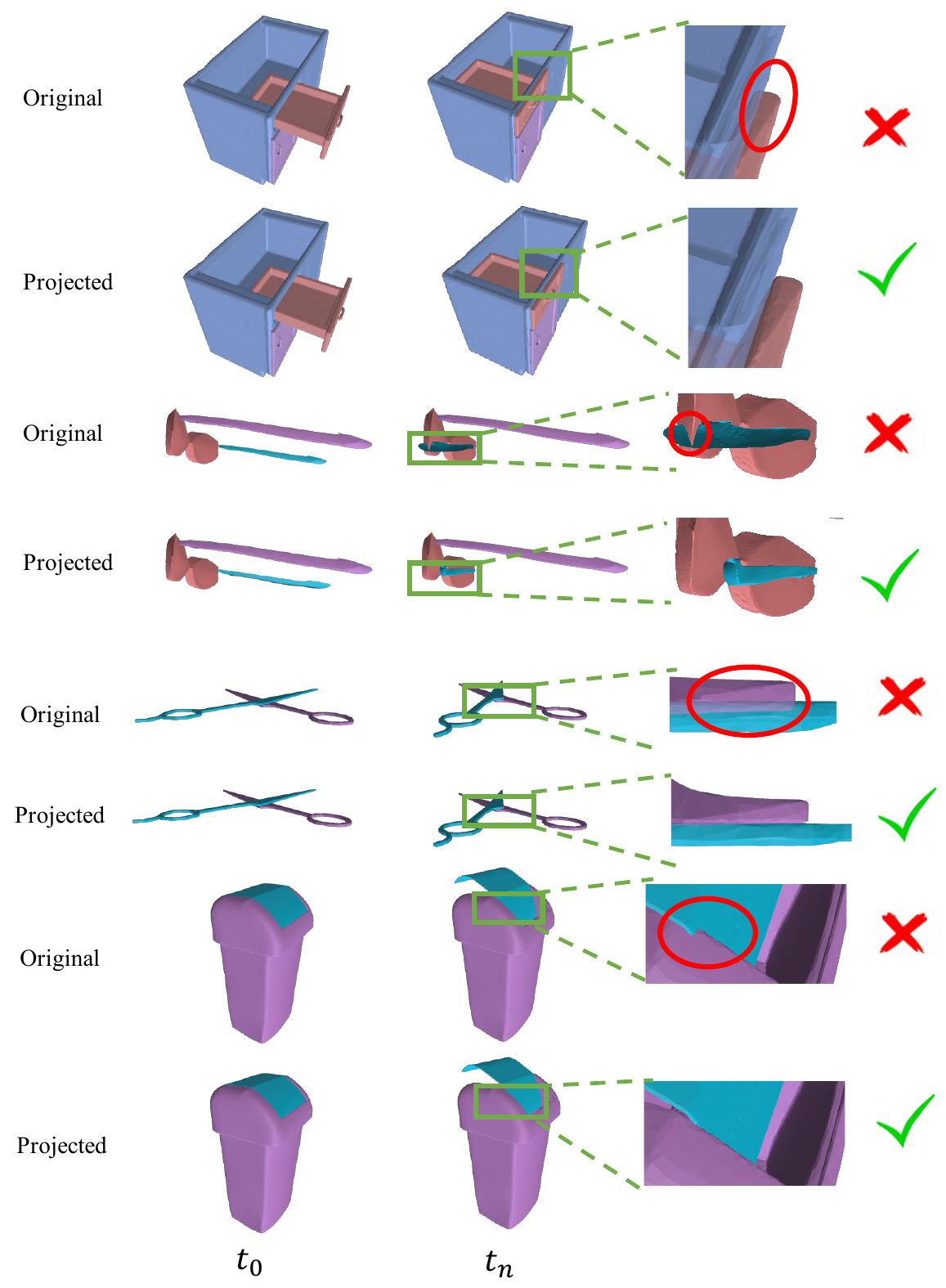}
    \caption{ \footnotesize
        \textbf{Visual evaluation on the effectiveness of the physics-aware projection strategy.}
        For every two lines, the upper line draws shapes without such correction while the second line draws corrected shapes. 
        }
        \vspace{-16pt}
    \label{fig_collision_projection_vis}
\end{figure}

\begin{figure}[ht]
    \centering
    \includegraphics[width=0.48\textwidth]{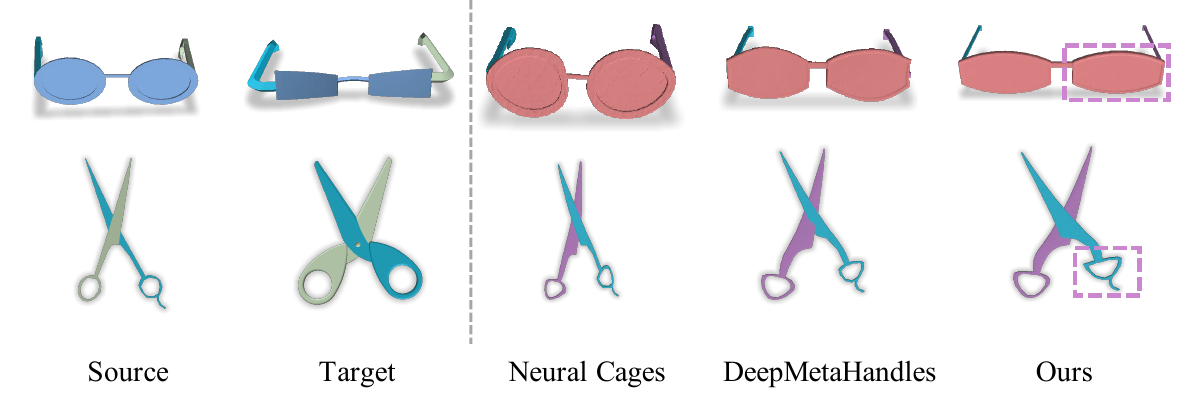}
    \caption{ \footnotesize
        Visual comparison on model's \textbf{target-driven deformation} ability. 
        }
        \vspace{-12pt}
    \label{fig_target_driven_def_comparison}
\end{figure}

\begin{figure}[ht]
    \centering
    \includegraphics[width=0.48\textwidth]{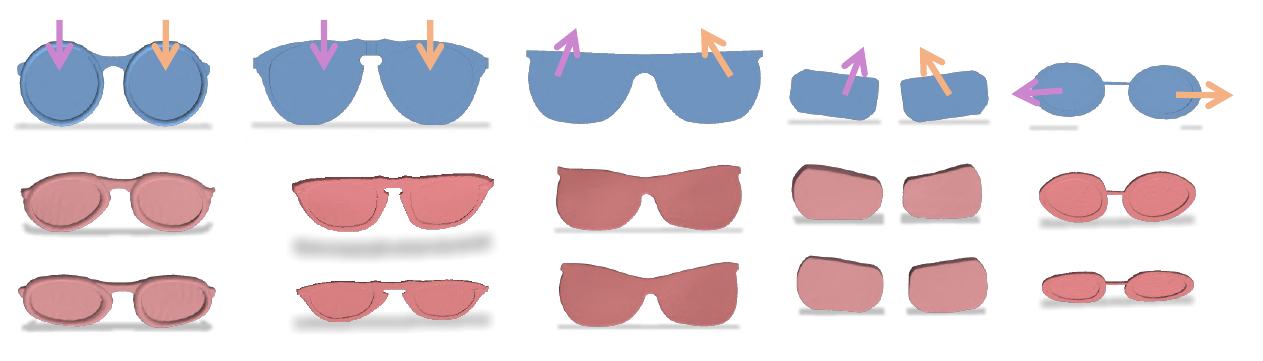}
    \vspace{-9pt}
    \caption{ \footnotesize
        \textbf{Visual evaluation on synchronized convex-level deformation bases.}
        The first line draws the template shape with deformation directions of synchronized deformation bases, while the following two lines are deformed shapes by their corresponding bases. Arrows are drawn to highlight the deformation direction. 
        }
        \vspace{-16pt}
    \label{fig_sync_basis}
\end{figure}



We evaluate our model on \nncat articulated object categories to test its few-shot generation ability for articulated objects. 

\vpara{Datasets.}
We evaluate our method on \nncat categories selected from PartNet-Mobility~\cite{xiang2020sapien} dataset following previous standard~\cite{li2020category}, namely Storage Furniture, Eyeglasses, Scissors, Oven, Lamp, and TrashCan. 
We select \nnpretrain instances from ShapeNet~\cite{chang2015shapenet} dataset, covering four categories: Table, Chair, Lamp, and Airplane for convex-level deformation pre-training. 
For each test category, we split it into a few-shot training set with 5 instances and a test set containing the remaining instances. 
For more details, please refer to the supplementary material. 

\vpara{Baselines.} 
We compare our method to PolyGen~\cite{nash2020polygen}, an auto-regressive style mesh generative model
and DeepMetaHandles~\cite{liu2021deepmetahandles}, a deformation-based mesh generative model.
To further adapt them for articulated object generation, we design a part-by-part generation approach for each of them and we defer details to the supp. 


\vpara{Metrics.}
We employ two kinds of metrics for evaluation: 
1) metrics for mesh generative models following previsou literature~\cite{liu2021deepmetahandles,luo2021diffusion,achlioptas2018learning}, 
that is the minimum matching distance (MMD), coverage (COV), 1-NN classifier accuracy (1-NNA), and Jenson-Shannon divergence (JSD)~\cite{yang2019pointflow}; and 
2) average penetration depth (APD) for physical validity evaluation. 
The MMD score evaluates the fidelity of the generated samples and COV detects mode collapse and measures the diversity of generated samples. 
The 1-NNA score is computed by testing the generated samples and the reference instances by a 1-NN classifier. 
We introduce it following~\cite{luo2021diffusion}. The classifier is not a network but classifies shapes into ``reference'' or ``training'' class based on the \textbf{N}earest \textbf{N}eighbour.
The JSD score computes the similarity between generated samples and reference samples. 
The APD score calculates the per-vertex average penetration depth averaged over all articulation simulation steps. 
We defer its details to the supp. 

\vpara{Experimental settings.} 
The number of projections are set to 5 and 10 at the training time and the test time respectively. 
The number of decomposed convexes may vary across categories and is detailed in the supp. 


\begin{table*}[t]
    \centering
    \caption{\footnotesize 
    \textbf{Quantitative evaluation.} Comparison between our method and baseline models on the few-shot articulated mesh generation task. 
    MMD is multiplied by $10^3$ and APD is multiplied by $10^2$.
    \textbf{Bold} numbers for best values. ``Avg.'' means ``Average Performance''. 
    } 
    \resizebox{0.8\linewidth}{!}{%
    \begin{tabular}{@{\;}c@{\;}|c|c|c|c|c|c|c|c@{\;}}
    \midrule
        \hline
        \specialrule{0em}{1pt}{0pt} 
        ~ & Method & \makecell[c]{Storage\\ Furniture} & Scissors &  Eyeglasses  & Oven & Lamp & TrashCan & Avg. \\ 
        \cline{1-9} 
        \specialrule{0em}{1pt}{0pt}
        
        \multirow{3}{*}{MMD~($\downarrow$)} & PolyGen~\cite{nash2020polygen} & 4.447 & 3.020 & 8.426 & 7.477 & 12.478 & 9.817 & 7.611
        \\ \cline{2-9} 
        \specialrule{0em}{1pt}{0pt}

        ~ & DeepMetaHandles~\cite{liu2021deepmetahandles} & \textbf{1.031} & 1.854 & 6.414 & 7.730 & 8.560 & 9.213 & 5.800
        \\ \cline{2-9} 
        \specialrule{0em}{1pt}{0pt}

        ~ & Ours & {1.058} & \textbf{1.495} & \textbf{6.062} & \textbf{7.009} & \textbf{7.133} & \textbf{8.430}  & \textbf{5.198} 
        \\ \cline{1-9} 
        \specialrule{0em}{1pt}{0pt}

        \multirow{3}{*}{COV~(\%, $\uparrow$)} & PolyGen~\cite{nash2020polygen} & 19.23 & 9.76 & 8.33 & 60.00  & 37.50 & 14.29 & 24.85 
        \\ \cline{2-9} 
        \specialrule{0em}{1pt}{0pt}

        ~ & DeepMetaHandles~\cite{liu2021deepmetahandles} & 43.93 & 24.63 & 15.00 & 60.00 & 50.00 & 17.14 & 35.12 
        \\ \cline{2-9} 
        \specialrule{0em}{1pt}{0pt}

        ~ & Ours & \textbf{75.33} & \textbf{57.89} & \textbf{29.82} & \textbf{60.00} & \textbf{62.50} & \textbf{17.14} & \textbf{50.45}
        \\ \cline{1-9} 
        \specialrule{0em}{1pt}{0pt}

        \multirow{3}{*}{1-NNA~(\%, $\downarrow$)} & PolyGen~\cite{nash2020polygen} & 99.46 & 98.28 & 98.71 & 98.04 & 99.22 & 92.68 & 97.73
        \\ \cline{2-9} 
        \specialrule{0em}{1pt}{0pt}

        ~ & DeepMetaHandles~\cite{liu2021deepmetahandles} & \textbf{97.72} & 98.07 & 98.33 & 98.50 & 94.65 & 86.05 & 95.55
        \\ \cline{2-9} 
        \specialrule{0em}{1pt}{0pt}

        ~ & Ours &  97.76 & \textbf{97.02}  & \textbf{98.26} & \textbf{96.59} & \textbf{92.44} & \textbf{72.09} & \textbf{92.36}  
        \\ \cline{1-9} 
        \specialrule{0em}{1pt}{0pt}

        \multirow{3}{*}{JSD~($\downarrow$)} & PolyGen~\cite{nash2020polygen} & 0.0791 & 0.2317 & 0.1350 & 0.2044 & 0.2761 & 0.2269 & 0.1922
        \\ \cline{2-9} 
        \specialrule{0em}{1pt}{0pt}

        ~ & DeepMetaHandles~\cite{liu2021deepmetahandles} & 0.0697 & 0.2277 & 0.0960 & 0.1768 & 0.2172 & 0.1881 & 0.1626
        \\ \cline{2-9} 
        \specialrule{0em}{1pt}{0pt}

        ~ & Ours &  \textbf{0.0290} & \textbf{0.1274} & \textbf{0.0681} & \textbf{0.1597} & \textbf{0.1874} & \textbf{0.0994} & \textbf{0.1118} 
        \\ \cline{1-9} 
        \specialrule{0em}{1pt}{0pt}

        \multirow{3}{*}{APD~($\downarrow$)} & PolyGen~\cite{nash2020polygen} & \textbf{0.1305} & \textbf{0.2592} & \textbf{0.0479} & {0.2548} & \textbf{5.323} & \textbf{0.0256} & \textbf{1.0068} 
        \\ \cline{2-9} 
        \specialrule{0em}{1pt}{0pt}


        ~ & DeepMetaHandles~\cite{liu2021deepmetahandles} &  0.2990 & 1.6670  & 0.3682 & 0.4408 & 6.3020 & 1.6961 & 1.7955 
        \\ \cline{2-9} 
        \specialrule{0em}{1pt}{0pt}

        ~ & Ours &  0.1700 & 1.3520  & 0.1707 & \textbf{0.1602} & 5.993 & 0.0693 & 1.3192 
        \\ \cline{1-9} 
        \specialrule{0em}{1pt}{0pt}

    \end{tabular}
    }
    \label{tb_exp_gen_comparison}
\end{table*}

\begin{table*}[htbp]
    \centering
    \caption{\footnotesize 
    \textbf{Ablation study} w.r.t. convex-level deformation transfer learning, hierarchical mesh generation, and physics-aware deformation correction. 
    For metrics of each ablated version, 
    we report their average value over all categories. 
    MMD is multiplied by $10^3$ and APD is multiplied by $10^2$. 
    \textbf{Bold} numbers for best values.  \emph{Italics} numbers for the second-best one. 
    } 
    \resizebox{0.85\linewidth}{!}{%
    \begin{tabular}{@{\;}c@{\;}|c|c|c|c|c|c@{\;}}
    \midrule
        \hline
        \specialrule{0em}{1pt}{0pt} 
        Ablation Type & Method & MMD~($\downarrow$) &  COV~(\%, $\uparrow$)  & 1-NNA~(\%, $\downarrow$) & JSD~($\downarrow$) & APD~($\downarrow$) \\ 
        \cline{1-7} 
        \specialrule{0em}{1pt}{0pt}
        

        \multirow{1}{*}{Hierarchical deformation} & Ours w/o Hier. & 7.170 & 36.41 & 95.43 & 0.1492 & 1.4964 
        \\ \cline{1-7} 
        \specialrule{0em}{1pt}{0pt}

        \multirow{3}{*}{\makecell[c]{Transfer learning \& \\ Fine-tuning}} & Ours w/o Transfer & 5.424 & 46.64 & 93.01 & 0.1159 & {1.3822}
        \\ \cline{2-7} 
        \specialrule{0em}{1pt}{0pt}

        ~ & Ours w/ Transfer (Half Data) & 5.201 & 49.43 & 92.81 & 0.1130 & \emph{1.3365}
        \\ \cline{2-7} 
        \specialrule{0em}{1pt}{0pt}

        ~ & Ours w/o Fine-tuning & 6.538 & 43.20 & 94.70 & 0.1437 & \emph{1.4530}
        \\ \cline{1-7} 
        \specialrule{0em}{1pt}{0pt}

        \multirow{3}{*}{\makecell[c]{Physics-aware \\ deformation correction}} & DeepMetaHandles w/ Phy. & 6.980 & 37.20 & 95.69 & 0.1587 & 1.5705
        \\ \cline{2-7} 
        \specialrule{0em}{1pt}{0pt}

        ~ & Ours w/o Phy. & \emph{5.211} & \textbf{50.83} & 92.57 & \emph{0.1060} & 1.8079
        \\ \cline{2-7} 
        \specialrule{0em}{1pt}{0pt}

        ~ & Ours w/o TTA & 5.214 & \emph{50.71} & \textbf{92.31} & \textbf{0.0992} & 1.6443
        \\ \cline{1-7} 
        \specialrule{0em}{1pt}{0pt}

        N/A & Ours & \textbf{5.198} & {50.45} & \emph{92.36} & {0.1118} & \textbf{1.3192}
        \\ \cline{1-7} 
        \specialrule{0em}{1pt}{0pt}

    \end{tabular}
    }
    \vspace{-10pt}
    \label{tb_exp_gen_comparison_abl}
\end{table*}

\vpara{Quantitative experimental evaluation: Few-shot articulated mesh generation.}
We summarize the 
quantitative evaluation results and comparisons to baseline methods on each articulated object category in Table~\ref{tb_exp_gen_comparison}. 
We can make the following observations: 
1) We can achieve better average performance on every metric than the baseline models. 
It demonstrates the power of our model to generate samples with better diversity, higher visual fidelity, and better physical validity than previous models from a small number of examples. 
2) On relatively rich categories (containing more than 30 instances) such as Scissors and Eyeglasses, our model can always outperform baseline methods by a large margin.
It indicates that our model can cover a wider distribution space than baseline methods by only training on a few examples. 
3) Our method can produce shapes with higher visual fidelity and better physical validity but not as a trade for diversity. By contrast, PolyGen generate samples that are more physically correct but exhibits very limited generative ability, \emph{i.e.,} poor COV and MMD scores. 

\vpara{Qualitative evaluation: Free deformation.}
In the free deformation setting, our model generates articulated meshes by deforming an input reference shape. 
It draw samples from the optimized object-level deformation coefficient distribution to deform input shapes. 
We draw deformed shapes from four representative categories, including Eyeglasses, Scissors, TrashCan, and Lamp in Figure~\ref{fig_free_gen_vis}. 
It demonstrates the ability of our model to create diverse variations by deforming input reference shapes.  
Compared with previous mesh deformation literature where the model always struggles to depict large geometry variations in the learned deformation space, our model mitigates this issue and is able to encode such deformations as observed in deformations of TrachCan bodies (line 3 of Figure~\ref{fig_free_gen_vis}). 
It mainly credits to our deformation coefficient distribution parameterization strategy, which is fit by discrete deformation coefficients, other than the uniform range adopted in~\cite{liu2021deepmetahandles}.

\vpara{Qualitative results: Target-driven deformation.}
We also conduct the target-driven experiments to demonstrate the superiority of our hierarchical deformation strategy over previous deformation literature, \emph{i.e.,} DeepMetaHandles~\cite{liu2021deepmetahandles}, and Neural Cages~\cite{yifan2020neural}. 
As shown in Figure~\ref{fig_target_driven_def_comparison}, our model can deform shapes to be more similar to their corresponding target shapes. It demonstrates the enhanced flexibility of our deformation strategy. 

\vpara{Convex deformation synchronization.}
We visualize the effectiveness of our convex deformation synchronization design by showing how synchronized deformation bases change shapes to produce plausible global mesh-level deformations in Figure~\ref{fig_sync_basis}. 
Besides, though not imposed directly, we do observe cross-instance similar deformation patterns, as also observed in \cite{liu2021deepmetahandles}. 

\vpara{Physics-aware projection.}
We compare objects synthesized by the network directly without and with physics-aware shape optimization in Figure~\ref{fig_collision_projection_vis}. 
Our shape optimization design can improve the physical validity of sampled shapes by 
resolving penetrations caused by either part translation (example 1 in Figure~\ref{fig_collision_projection_vis}) or revolution (example 2,3,4), in either the body part (example 1,2) or around the joint (example 3,4). 

\section{Ablation Study}
\vpara{Transfer learning and fine-tuning for the convex-level deformation module.}
In the few-shot generation design, the transfer learning technique plays an important role in enriching deformation space of the target category. Meanwhile, the fine-tuning process benefits the quality and diversity by learning category-specific deformation patterns. 
Our further analysis demonstrates that 1) 
The transfer learning's power can be boosted by increasing the amount of source data and is related to the affinity between source and target categories; 
2) The fine-tuning process is crucial for us to maintain high quality while achieving high diversity. 
We create three ablated models by ablating the transferring learning (``Ours w/o Transfer''), using half amount of the original data for transferring (``Ours w/ Transfer (Half Data)''), and ablating the fine-tuning process (``Ours w/o Fine-tuning'') and test their performance. Observations in Table~\ref{tb_exp_gen_comparison_abl} can validate the importance of the transfer learning and the fine-tuning process. 

\vpara{Hierarchical mesh deformation.}
We adopt the divide-and-conquer philosophy and design a hierarchical mesh deformation strategy to learn a diverse mesh deformation space. 
To demonstrate its superiority over simple part-level deformation and composition, we ablate such design and treat 
parts as the leaf deformation units (denoted as ``Ours w/o Hier.''). 
As shown in Table~\ref{tb_exp_gen_comparison_abl}, this way we observe immediate dropping of the performance on all metrics measuring the generative ability. 
This further evidence the value of our fine-grained decomposition and the hierarchical deformation space learning. 

\vpara{Physics-aware deformation correction.}
We design $\mathcal{L}_{phy}$ and $\mathcal{L}_{proj}$ to provide physical supervision and perform collision response-based shape optimization respectively. 
This way we are able to improve the physical validity of shapes deformed by our framework. 
To further validate them as solid contributions and versatile strategies not only work for our method, we create the following variants and test their performance: 1) ``Ours w/o Phy.'' by ablating both the shape optimization and the physical supervision, 2) ``Ours w/o TTA''  by only ablating the shape optimization strategy at the test time, 
and 3) ``DeepMetaHandles w/ Phy'' by integrating such two designs into the baseline DeepMetaHandles's framework. 
From Table~\ref{tb_exp_gen_comparison_abl}, we can make the following observations: 1) Our physics-aware correction strategy is a versatile design that can be easily integrated into another deformation-based mesh generative model, improving its performance effectively; 2) Only training-time physics-aware corrections can improve physical-related performance by guiding the convex-level deformation module stably and effectively; 3) Further imposing test time optimizations is important for us to arrive at high-quality samples finally. 

\section{Conclusion and Limitations}
We tackle the few-shot articulated mesh generation problem with 1) a hierarchical deformation model with transfer learning; and 2) a deformation correction scheme. 

\noindent \textbf{Limitations.} Currently, our work is limited to a category-level setting with the articulation chain and the range of articulation states assumed known. Developing a generation method without such assumption would increase its practical value and is an interesting future research direction. Besides, the deformation correction scheme relies on hand-crafted chain of articulation states to detect self-penetrations and optimize shapes based on that. A natural alternative can be detecting articulation states from real-world images. Moreover, the quality of our generated results are restricted by that of the training data. A smart self-correction strategy, beyond mitigating self-penetrations only, may be designed to improve the validity. 

\clearpage
{\small
\bibliographystyle{ieee_fullname}
\bibliography{egbib}

\begin{thebibliography}{10}\itemsep=-1pt

\bibitem{achlioptas2018learning}
Panos Achlioptas, Olga Diamanti, Ioannis Mitliagkas, and Leonidas Guibas.
\newblock Learning representations and generative models for 3d point clouds.
\newblock In {\em International conference on machine learning}, pages 40--49.
  PMLR, 2018.

\bibitem{antoniou2017data}
Antreas Antoniou, Amos Storkey, and Harrison Edwards.
\newblock Data augmentation generative adversarial networks.
\newblock {\em arXiv preprint arXiv:1711.04340}, 2017.

\bibitem{bartunov2018few}
Sergey Bartunov and Dmitry Vetrov.
\newblock Few-shot generative modelling with generative matching networks.
\newblock In {\em International Conference on Artificial Intelligence and
  Statistics}, pages 670--678. PMLR, 2018.

\bibitem{chang2015shapenet}
Angel~X Chang, Thomas Funkhouser, Leonidas Guibas, Pat Hanrahan, Qixing Huang,
  Zimo Li, Silvio Savarese, Manolis Savva, Shuran Song, Hao Su, et~al.
\newblock Shapenet: An information-rich 3d model repository.
\newblock {\em arXiv preprint arXiv:1512.03012}, 2015.

\bibitem{chen2020bsp}
Zhiqin Chen, Andrea Tagliasacchi, and Hao Zhang.
\newblock Bsp-net: Generating compact meshes via binary space partitioning.
\newblock In {\em Proceedings of the IEEE/CVF Conference on Computer Vision and
  Pattern Recognition}, pages 45--54, 2020.

\bibitem{chou2022diffusionsdf}
Gene Chou, Yuval Bahat, and Felix Heide.
\newblock Diffusionsdf: Conditional generative modeling of signed distance
  functions.
\newblock {\em arXiv preprint arXiv:2211.13757}, 2022.

\bibitem{de2018end}
Filipe de Avila Belbute-Peres, Kevin Smith, Kelsey Allen, Josh Tenenbaum, and
  J~Zico Kolter.
\newblock End-to-end differentiable physics for learning and control.
\newblock {\em Advances in neural information processing systems}, 31, 2018.

\bibitem{gao2022get3d}
Jun Gao, Tianchang Shen, Zian Wang, Wenzheng Chen, Kangxue Yin, Daiqing Li, Or
  Litany, Zan Gojcic, and Sanja Fidler.
\newblock Get3d: A generative model of high quality 3d textured shapes learned
  from images.
\newblock {\em arXiv preprint arXiv:2209.11163}, 2022.

\bibitem{garcia2013cages}
Francisco~Gonz{\'a}lez Garc{\'\i}a, Teresa Paradinas, Narcis Coll, and Gustavo
  Patow.
\newblock * cages: a multilevel, multi-cage-based system for mesh deformation.
\newblock {\em ACM Transactions on Graphics (TOG)}, 32(3):1--13, 2013.

\bibitem{gu2021lofgan}
Zheng Gu, Wenbin Li, Jing Huo, Lei Wang, and Yang Gao.
\newblock Lofgan: Fusing local representations for few-shot image generation.
\newblock In {\em Proceedings of the IEEE/CVF International Conference on
  Computer Vision}, pages 8463--8471, 2021.

\bibitem{hong2022fixing}
Yining Hong, Kaichun Mo, Li Yi, Leonidas~J Guibas, Antonio Torralba, Joshua~B
  Tenenbaum, and Chuang Gan.
\newblock Fixing malfunctional objects with learned physical simulation and
  functional prediction.
\newblock In {\em Proceedings of the IEEE/CVF Conference on Computer Vision and
  Pattern Recognition}, pages 1413--1423, 2022.

\bibitem{hong2020matchinggan}
Yan Hong, Li Niu, Jianfu Zhang, and Liqing Zhang.
\newblock Matchinggan: Matching-based few-shot image generation.
\newblock In {\em 2020 IEEE International Conference on Multimedia and Expo
  (ICME)}, pages 1--6. IEEE, 2020.

\bibitem{hong2022deltagan}
Yan Hong, Li Niu, Jianfu Zhang, and Liqing Zhang.
\newblock Deltagan: Towards diverse few-shot image generation with
  sample-specific delta.
\newblock In {\em European Conference on Computer Vision}, pages 259--276.
  Springer, 2022.

\bibitem{hong2020f2gan}
Yan Hong, Li Niu, Jianfu Zhang, Weijie Zhao, Chen Fu, and Liqing Zhang.
\newblock F2gan: Fusing-and-filling gan for few-shot image generation.
\newblock In {\em Proceedings of the 28th ACM international conference on
  multimedia}, pages 2535--2543, 2020.

\bibitem{Hu2022physicalinteraction}
Haoyu Hu, Xinyu Yi, Hao Zhang, Jun-Hai Yong, and Feng Xu.
\newblock Physical interaction: Reconstructing hand-object interactions with
  physics.
\newblock In {\em {SIGGRAPH} Asia 2022 Conference Papers}. {ACM}, nov 2022.

\bibitem{hu2019difftaichi}
Yuanming Hu, Luke Anderson, Tzu-Mao Li, Qi Sun, Nathan Carr, Jonathan
  Ragan-Kelley, and Fr{\'e}do Durand.
\newblock Difftaichi: Differentiable programming for physical simulation.
\newblock {\em arXiv preprint arXiv:1910.00935}, 2019.

\bibitem{hu2019chainqueen}
Yuanming Hu, Jiancheng Liu, Andrew Spielberg, Joshua~B Tenenbaum, William~T
  Freeman, Jiajun Wu, Daniela Rus, and Wojciech Matusik.
\newblock Chainqueen: A real-time differentiable physical simulator for soft
  robotics.
\newblock In {\em 2019 International conference on robotics and automation
  (ICRA)}, pages 6265--6271. IEEE, 2019.

\bibitem{lan2022dream}
Ke Lan.
\newblock Dream fusion in octahedral spherical hohlraum.
\newblock {\em Matter and Radiation at Extremes}, 7(5):055701, 2022.

\bibitem{li2017grass}
Jun Li, Kai Xu, Siddhartha Chaudhuri, Ersin Yumer, Hao Zhang, and Leonidas
  Guibas.
\newblock Grass: Generative recursive autoencoders for shape structures.
\newblock {\em ACM Transactions on Graphics (TOG)}, 36(4):1--14, 2017.

\bibitem{li2022diffusion}
Muheng Li, Yueqi Duan, Jie Zhou, and Jiwen Lu.
\newblock Diffusion-sdf: Text-to-shape via voxelized diffusion.
\newblock {\em arXiv preprint arXiv:2212.03293}, 2022.

\bibitem{li2020category}
Xiaolong Li, He Wang, Li Yi, Leonidas~J Guibas, A~Lynn Abbott, and Shuran Song.
\newblock Category-level articulated object pose estimation.
\newblock In {\em Proceedings of the IEEE/CVF conference on computer vision and
  pattern recognition}, pages 3706--3715, 2020.

\bibitem{liu2021deepmetahandles}
Minghua Liu, Minhyuk Sung, Radomir Mech, and Hao Su.
\newblock Deepmetahandles: Learning deformation meta-handles of 3d meshes with
  biharmonic coordinates.
\newblock {\em arXiv preprint arXiv:2102.09105}, 2021.

\bibitem{liu2023self}
Xueyi Liu, Ji Zhang, Ruizhen Hu, Haibin Huang, He Wang, and Li Yi.
\newblock Self-supervised category-level articulated object pose estimation
  with part-level se (3) equivariance.
\newblock In {\em The Eleventh International Conference on Learning
  Representations}, 2023.

\bibitem{luo2021diffusion}
Shitong Luo and Wei Hu.
\newblock Diffusion probabilistic models for 3d point cloud generation.
\newblock In {\em Proceedings of the IEEE/CVF Conference on Computer Vision and
  Pattern Recognition}, pages 2837--2845, 2021.

\bibitem{mezghanni2022physical}
Mariem Mezghanni, Th{\'e}o Bodrito, Malika Boulkenafed, and Maks Ovsjanikov.
\newblock Physical simulation layer for accurate 3d modeling.
\newblock In {\em Proceedings of the IEEE/CVF Conference on Computer Vision and
  Pattern Recognition}, pages 13514--13523, 2022.

\bibitem{mezghanni2021physically}
Mariem Mezghanni, Malika Boulkenafed, Andre Lieutier, and Maks Ovsjanikov.
\newblock Physically-aware generative network for 3d shape modeling.
\newblock In {\em Proceedings of the IEEE/CVF Conference on Computer Vision and
  Pattern Recognition}, pages 9330--9341, 2021.

\bibitem{nash2020polygen}
Charlie Nash, Yaroslav Ganin, SM~Ali Eslami, and Peter Battaglia.
\newblock Polygen: An autoregressive generative model of 3d meshes.
\newblock In {\em International conference on machine learning}, pages
  7220--7229. PMLR, 2020.

\bibitem{peng2021shape}
Songyou Peng, Chiyu Jiang, Yiyi Liao, Michael Niemeyer, Marc Pollefeys, and
  Andreas Geiger.
\newblock Shape as points: A differentiable poisson solver.
\newblock {\em Advances in Neural Information Processing Systems},
  34:13032--13044, 2021.

\bibitem{ren2023diffmimic}
Jiawei Ren, Cunjun Yu, Siwei Chen, Xiao Ma, Liang Pan, and Ziwei Liu.
\newblock Diffmimic: Efficient motion mimicking with differentiable physics.
\newblock 2023.

\bibitem{rombach2022high}
Robin Rombach, Andreas Blattmann, Dominik Lorenz, Patrick Esser, and Bj{\"o}rn
  Ommer.
\newblock High-resolution image synthesis with latent diffusion models.
\newblock In {\em Proceedings of the IEEE/CVF Conference on Computer Vision and
  Pattern Recognition}, pages 10684--10695, 2022.

\bibitem{saharia2022photorealistic}
Chitwan Saharia, William Chan, Saurabh Saxena, Lala Li, Jay Whang, Emily
  Denton, Seyed Kamyar~Seyed Ghasemipour, Burcu~Karagol Ayan, S~Sara Mahdavi,
  Rapha~Gontijo Lopes, et~al.
\newblock Photorealistic text-to-image diffusion models with deep language
  understanding.
\newblock {\em arXiv preprint arXiv:2205.11487}, 2022.

\bibitem{shen2021deep}
Tianchang Shen, Jun Gao, Kangxue Yin, Ming-Yu Liu, and Sanja Fidler.
\newblock Deep marching tetrahedra: a hybrid representation for high-resolution
  3d shape synthesis.
\newblock {\em Advances in Neural Information Processing Systems},
  34:6087--6101, 2021.

\bibitem{shu20203d}
Dule Shu, James Cunningham, Gary Stump, Simon~W Miller, Michael~A Yukish,
  Timothy~W Simpson, and Conrad~S Tucker.
\newblock 3d design using generative adversarial networks and physics-based
  validation.
\newblock {\em Journal of Mechanical Design}, 142(7):071701, 2020.

\bibitem{shue20223d}
J~Ryan Shue, Eric~Ryan Chan, Ryan Po, Zachary Ankner, Jiajun Wu, and Gordon
  Wetzstein.
\newblock 3d neural field generation using triplane diffusion.
\newblock {\em arXiv preprint arXiv:2211.16677}, 2022.

\bibitem{singer2022make}
Uriel Singer, Adam Polyak, Thomas Hayes, Xi Yin, Jie An, Songyang Zhang, Qiyuan
  Hu, Harry Yang, Oron Ashual, Oran Gafni, et~al.
\newblock Make-a-video: Text-to-video generation without text-video data.
\newblock {\em arXiv preprint arXiv:2209.14792}, 2022.

\bibitem{sung2020deformsyncnet}
Minhyuk Sung, Zhenyu Jiang, Panos Achlioptas, Niloy~J Mitra, and Leonidas~J
  Guibas.
\newblock Deformsyncnet: Deformation transfer via synchronized shape
  deformation spaces.
\newblock {\em arXiv preprint arXiv:2009.01456}, 2020.

\bibitem{wen2019pixel2mesh}
Chao Wen, Yinda Zhang, Zhuwen Li, and Yanwei Fu.
\newblock Pixel2mesh++: Multi-view 3d mesh generation via deformation.
\newblock In {\em Proceedings of the IEEE/CVF international conference on
  computer vision}, pages 1042--1051, 2019.

\bibitem{weng2021captra}
Yijia Weng, He Wang, Qiang Zhou, Yuzhe Qin, Yueqi Duan, Qingnan Fan, Baoquan
  Chen, Hao Su, and Leonidas~J Guibas.
\newblock Captra: Category-level pose tracking for rigid and articulated
  objects from point clouds.
\newblock In {\em Proceedings of the IEEE/CVF International Conference on
  Computer Vision}, pages 13209--13218, 2021.

\bibitem{xiang2020sapien}
Fanbo Xiang, Yuzhe Qin, Kaichun Mo, Yikuan Xia, Hao Zhu, Fangchen Liu, Minghua
  Liu, Hanxiao Jiang, Yifu Yuan, He Wang, et~al.
\newblock Sapien: A simulated part-based interactive environment.
\newblock In {\em Proceedings of the IEEE/CVF Conference on Computer Vision and
  Pattern Recognition}, pages 11097--11107, 2020.

\bibitem{xu2021umpnet}
Zhenjia Xu, Zhanpeng He, and Shuran Song.
\newblock Umpnet: Universal manipulation policy network for articulated
  objects.
\newblock {\em arXiv preprint arXiv:2109.05668}, 2021.

\bibitem{yang2019pointflow}
Guandao Yang, Xun Huang, Zekun Hao, Ming-Yu Liu, Serge Belongie, and Bharath
  Hariharan.
\newblock Pointflow: 3d point cloud generation with continuous normalizing
  flows.
\newblock In {\em Proceedings of the IEEE/CVF International Conference on
  Computer Vision}, pages 4541--4550, 2019.

\bibitem{yifan2020neural}
Wang Yifan, Noam Aigerman, Vladimir~G Kim, Siddhartha Chaudhuri, and Olga
  Sorkine-Hornung.
\newblock Neural cages for detail-preserving 3d deformations.
\newblock In {\em Proceedings of the IEEE/CVF Conference on Computer Vision and
  Pattern Recognition}, pages 75--83, 2020.

\bibitem{zeng2022lion}
Xiaohui Zeng, Arash Vahdat, Francis Williams, Zan Gojcic, Or Litany, Sanja
  Fidler, and Karsten Kreis.
\newblock Lion: Latent point diffusion models for 3d shape generation.
\newblock {\em arXiv preprint arXiv:2210.06978}, 2022.

\bibitem{zhou2021pvd}
Linqi Zhou, Yilun Du, and Jiajun Wu.
\newblock 3d shape generation and completion through point-voxel diffusion.
\newblock In {\em Proceedings of the IEEE/CVF International Conference on
  Computer Vision}, pages 5826--5835, 2021.

\end{thebibliography}
}

\clearpage
\appendix

The appendix provides a list of supplemental materials to support the main paper.
\begin{itemize}
    \item \textbf{Further Explanations on the Method -- }   We provide additional details for some components and algorithms to complement the main paper.
        \begin{itemize}
            \item \emph{Convex-Level Deformation Generative Model} (Sec.~\ref{sec_supp_cvx_def}). 
            We include a more detailed explanation regarding the model including two tricks we use to parameterize deformations
            \item \emph{Convex Deformation Synchronization} (Sec.~\ref{sec_supp_cvx_synchronization}). 
            We explain our alternative optimization strategy to calculate synchronization matrices in more detail.
            \item \emph{Physics-Aware Deformation Correction} (Sec.~\ref{sec_supp_part_composition}). 
            We explain the calculation processes of the physical penalty $\mathcal{L}_{phy}$ and the projection loss $\mathcal{L}_{proj}$ in more detail. 
            \item \emph{Additional Explanations} (Sec.~\ref{sec_supp_method_additional_details}). 
            We include more details regarding the practical implementation of the method. 
        \end{itemize}
    \item \textbf{Additional Experiments -- } We present additional experiments to further prove the effectiveness of our method.
        \begin{itemize}
            \item \emph{Few-Shot Mesh Generation} (Sec.~\ref{sec_supp_few_shot_mesh_gen}).  We further examine the few-shot generation ability of our hierarchical deformation strategy by testing it on different few-shot generation settings and also on rigid mesh categories. 
            \item \emph{Transfer Learning for Convex Deformations} (Sec.~\ref{sec_supp_tranfer_cvx_def}). 
            We demonstrate why we choose to use convexes as intermediates to transfer cross-category shared shape patterns. 
            Besides, we also explore the influence of source categories in the transfer learning on the few-shot generation performance.  
            \item \emph{Generation via Deformation} (Sec.~\ref{sec_supp_generation_via_deformation}). We leverage a different mesh generation technique and design a second approach trying to solve the problem. 
            By comparing this strategy to our method, we cast some thoughts on the design philosophy of strategies to solve the few-shot and physically-aware generation challenges. 
        \end{itemize}
    \item \textbf{Experimental Settings -- }   
    We provide additional information about our experimental settings. 
        \begin{itemize}
            \item \emph{Datasets} (Sec.~\ref{sec_supp_datasets}). We provide more information on our datasets for pre-training and evaluation. 
            \item \emph{Baselines} (Sec.~\ref{sec_supp_baselines_imple}).  We explain our modifications and improvements on baseline methods so that we can adapt them to the articulated mesh generation problem. 
            \item \emph{Metrics} (Sec.~\ref{sec_supp_metrics}).  We provide additional calculation details of the evaluation metrics. 
            \item \emph{Additional Experimental Settings} (Sec.~\ref{sec_supp_addtional_imple_details}). We further discuss some additional experimental settings. 
        \end{itemize}
\end{itemize}



\section{Further Explanations on the Method} \label{sec_method_details_supp}

\subsection{Convex-Level Generative Model} \label{sec_supp_cvx_def}

In the method, we design a convex-level generative model to parameterize vertex-level deformation offset $d_c$ into a low dimensional space. 
We leverage two tricks to parameterize $d_c$: 1) using cages to control per-vertex deformations and 2) using dictionaries to record common deformation patterns. 
We elaborate on details of the above tricks that are not covered in the main text in the following text. 

\vpara{Cages to control convex deformation. }
To form the cage $t_c$ of the convex $c$ containing $N_c$ vertices, we deform a template mesh based upon the shape of the convex $c$. 
Specifically, given a template mesh, \emph{i.e.,} a sphere surface mesh, $t_c$ with $N_t (N_t \ll  N_c)$ vertices, we deform $t_c$ to form the cage of $c$ via the following steps: 
1) Assume the vertex sets of $t_c$ and $c$ are $\mathcal{V}_t$ and $\mathcal{V}_c$ respectively. 
Find a mapping from each cage vertex, saying $v_t \in \mathcal{V}_t$, to a vertex $v_c$ in the convex $c$, \emph{i.e.}, 
$m(v_t) = v_c, v_t \in \mathcal{V}_t, v_c \in \mathcal{V}_c (m(v_{t_1}) \neq m(v_{t_2}), \forall v_{t_1} \neq v_{t_2})$
such that we can minimize $\sum_{v_t \in \mathcal{V}_t} \Vert m(v_t) - v_t\Vert_2$. We use ``linear\_sum\_assignment'' function implemented in package ``scipy'' to find the mapping $m(\cdot)$. 
2) Deform $v_t$ to $\hat{v}_t = v_t + (1 - \epsilon) \cdot (m(v_t) - v_t)$, where $\epsilon$ is a hyper-parameter, which is set to 0.05 in our experiment.

Such a heuristic deformation strategy works well in our problem considering shapes we want to deform here are near-to-convex segments. 

\vpara{Using dictionaries to record common deformation patterns.} 
We record common cage deformation patterns using deformation bases.
Following a previous work which also learns deformation bases to represent common deformation patterns~\cite{liu2021deepmetahandles}, we wish the deformation basses predicted by our network for each cage should be able to cover the entire deformation space such that each possible cage deformation can be decomposed into a linear combination of these bases. 
It is encouraged by our learning objective guided via the convex deformation loss $\mathcal{L}_C$ that minimizes the Chamfer Distance between each deformed convex and the target convex.
The Chamfer Distance between two convexes is defined on 4096 points sampled from their surfaces. 

At the same time, we wish deformation bases predicted by our network to have the following two properties: 
1) Deformation bases should be relatively orthogonal to each other to avoid recording redundant deformation patterns and to cover independent deformation patterns. 
2) Deformation bases should be in a low dimensional space, activating as few vertices as possible. 
Therefore, we further add an orthogonal loss $\mathcal{L}_{orth}$ and a sparse loss $\mathcal{L}_{sp}$ for regularization purpose, following~\cite{liu2021deepmetahandles}. 
Among them, $\mathcal{L}_{orth}$ penalizes ``dot products'' between different deformation bases. 
And $\mathcal{L}_{sp}$ penalizes the $l1$-norm of each deformation basis. 
Such two penalties are added as additional regularization to optimize the hierarchical deformation-based generative model together with $\mathcal{L}_{C}$ and the physical penalty term $\mathcal{L}_{phy}$. 

\subsection{Convex Deformation Synchronization} \label{sec_supp_cvx_synchronization}
After learning the conditional generative model for each convex $c$, we further design a convex deformation synchronization strategy to compose all the individual convex deformation spaces to the whole mesh-level deformation space. 
Given a set of articulated object meshes $\mathcal{A}$ from a certain category and an articulated mesh $a\in \mathcal{A}$, assuming the mesh $a$ is segmented into $M$ convexes and each convex is equipped with a deformation model $g_{\mathcal{C}}(\mathbf{z}_{c_m}|c_m)$, our goal is to replace $\mathbf{z}_{c_m}$ with $S_{c_m}\mathbf{z}$ so that sampling the shared noise parameter $\mathbf{z}$ results in a globally consistent mesh deformation. 
To compute the synchronization transformation $S_{c_m}$, we consider the deformation from $a$ to other articulated meshes $a^i\in\mathcal{A}$. 
In particular, for each $a^i$, we optimize for a set of deformation coefficients $\{\mathbf{y}_m^i\}$ so that each convex $c_m$ in mesh $a$ could deform into the corresponding convex $c_m^i$ in mesh $a^i$ following the deformation model $g_{\mathcal{C}}(\mathbf{z}_{c_m}|c_m,\mathbf{z}_{c_m}=\mathbf{y}_m^i)$. 
We can then estimate the synchronization transformations $\{S_{c_m}\}$ by solving the following optimization problem:
\begin{align}
    \underset{\{S_{c_m}\}, \{\mathbf{z}^i\}}{\text{minimize}} \sum_{i=1}^{\vert\mathcal{A}\vert}\sum_{m=1}^M\Vert B_{c_m}S_{c_m}\mathbf{z}^i-B_{c_m}\mathbf{y}_m^i\Vert_2,
    \label{eq_sync_supp}
\end{align} 
where $B_{c_m}$ is the deformation bases of convex $c_m$ and $\mathbf{z}^i$ is a global deformation coefficient from mesh $a$ to $a^i$ shared across all convexes. 
We solve the above optimization problem via alternatively optimizing the synchronization transformations $\{S_{c_m}\}$ and the global deformation coefficients $\{\mathbf{z}^i\}$. 

Specifically, we optimize equation~\ref{eq_sync_supp} by alternatively taking the following two steps: 
\begin{itemize}
    \item Fix $\{ S_{c_m} \}$, optimize each global deformation coefficient $\mathbf{z}^i$ from $a$ to $a^i$ via the global deformation coefficients optimization algorithm~\ref{algo_sync_opt_global_coeffs}.
    The algorithm takes the convex deformation bases $\{ B_{c_m} \}$, current synchronization transformations $\{ S_{c_m} \}$, and convex deformation coefficients $\{ \mathbf{y}_m^i \}$ as input, and outputs the optimized $\mathbf{z}^i$. 
    \item Fix $\{  \mathbf{z}^i \}$,  optimize each synchronization transformation $S_{c_m}$ for each convex $c_m$ via the synchronization transformation matrices optimization algorithm~\ref{algo_sync_opt_sync_matrix}. 
    It takes the  convex deformation bases $\{ B_{c_m} \}$, current global deformation coefficients $\{ \mathbf{z}^i \}$, and convex deformation coefficients $\{ \mathbf{y}_m^i \}$ as input, and output the optimized $S_{c_m}$. 
\end{itemize}

By looping the above two optimization steps several times (\emph{i.e.}, 100 in our implementation), we finally get the optimized synchronization transformation matrices $\{ S_{c_m} \}$ and global deformation coefficient $\{ \mathbf{z}^i \}$. 
Then the distribution of the global shape deformation coefficient $\mathbf{z}$ is modeled by a mixture of Gaussian fit to the optimized $\{ \mathbf{z}^i \}$. 

Please note that the above approach (Algorithm~\ref{algo_sync_opt_sync_matrix},~\ref{algo_sync_opt_global_coeffs}) is an approximate solution and is not affected by $\{ B_{c_m} \}$. 
However, $\{ B_{c_m} \}$ indeed influence the optimization objective outlined in Eq. 2 in the main text and cannot be omitted. 

\begin{algorithm}[H]
\small 
\caption{\textbf{Synchronization transformation matrices optimization}. 
}
\label{algo_sync_opt_sync_matrix}
\footnotesize
    \begin{algorithmic}[1]
        \Require
            Deformation bases for each convex $\{  B_{c_m}\}$. 
            Global deformation coefficients $\{ \mathbf{z}^i \}$ from $a$ to other articulated meshes $\{ a^i \}$. 
            Deformation coefficients $\{ \mathbf{y}_m^i \}$ from each convex $c_m$ to the corresponding convex of the articulated mesh $a^i$. 
        \Ensure
            Synchronization transformation matrix $S_{c_m}$ of the convex $c_m$.

        \State $\mathbf{Z} \leftarrow \text{Stack}(\{ \mathbf{z}^i \})$
        \State $\mathbf{Y}_m \leftarrow \text{Stack}( \{ \mathbf{y}_m^i \} )$
        \State $[\mathbf{U}, \mathbf{\Sigma}, \mathbf{V}^T] \leftarrow \text{SVD}(\mathbf{Z})$
        \State $[\mathbf{U}_m, \mathbf{\Sigma}_m, \mathbf{V}_m^T] \leftarrow \text{SVD}(\mathbf{Y}_m)$
        \State ${S}_{c_m} \leftarrow \mathbf{U}_m \mathbf{\Sigma}_m \mathbf{V}_m^T \mathbf{V} \mathbf{\Sigma}^+ \mathbf{U}^T$ \\
        \Return ${S}_{c_m}$
    \end{algorithmic}
\end{algorithm}

\begin{algorithm}[H]
\small 
\caption{\textbf{Global deformation coefficients optimization}. 
``lsq'' denotes the least square solver. 
}
\label{algo_sync_opt_global_coeffs}
\footnotesize
    \begin{algorithmic}[1]
        \Require
            Deformation bases for each convex $\{  B_{c_m}\}$. 
            Synchronization transformations $\{ S_{c_m} \}$. 
            Deformation coefficients $\mathbf{y}_m^i$ from each convex $c_m$ to the corresponding convex of the articulated mesh $a^i$. 
        \Ensure
            Global deformation coefficients $\mathbf{z}^i$ from $a$ to $a^i$. 

        \State $\mathcal{S}_{\mathbf{z}^i} \leftarrow \emptyset$
        \For{$m = 1$ to $M$}
            \State $\hat{\mathbf{z}}_m^i\leftarrow \text{lsq}({S}_{c_m}, \mathbf{z}_m^i)$
            \State $\mathcal{S}_{\mathbf{z}^i} \leftarrow \mathcal{S}_{\mathbf{z}^i}  \cup \{ \hat{\mathbf{z}}_m^i \}$
        \EndFor
        \State $\mathbf{z}^i = \text{Average}(\mathcal{S}_{\mathbf{z}^i})$ \\
        \Return $\mathbf{z}^i$
    \end{algorithmic}
\end{algorithm}

\subsection{Physics-Aware Deformation Correction} \label{sec_supp_part_composition}
In our method, we further add a physics-aware deformation correction scheme to 1) encourage the hierarchical deformation model to generate physically-realistic deformations and 2) optimize synthesized articulated meshes such that they can support correct articulation functions. 

We leverage physical simulation and a collision response-based shape optimization strategy to realize this vision. 
Two losses are involved in the correction strategy: 1) a physical penalty term $\mathcal{L}_{phy}$ measuring self-penetrations and 2) a projection loss  $\mathcal{L}_{proj}$ guiding how to project penetrated vertices to resolve observed penetrations. 

We implement $\mathcal{L}_{phy}$  and $\mathcal{L}_{proj}$ manually with no simulators. 
Given an articulated mesh $a$, we illustrate the details of their computing process in Algorithms~\ref{algo_phy_sim_single_part_single_sim}~\ref{algo_phy_sim}. 

For each category, the $K$ articulation states is formed by $C$ independent articulation chains. Each articulation chain consists of a moving part $p_{mov}$ and other static parts in their specific articulation states. $p_{mov}$ is articulated through the whole articulation range when articulating the object by the articulation chain. However, the valid articulated states for resting parts may vary across different categories. For eyeglasses, when articulating one leg, the other one should be put into 0 degree or 90 degree.

\begin{algorithm}[H]
\small 
\caption{\textbf{Single simulation}. 
Single-part articulation simulation losses. 
``NONE\_MOTION'' indicates fixed parts. 
}
\label{algo_phy_sim_single_part_single_sim}
\footnotesize
    \begin{algorithmic}[1]
        \Require
            Part mesh 
            $p_{mov} = (\mathbf{V}_{mov}, \mathbf{E}_{mov})$ to articulate at the rest pose; 
            Convex mesh $p_{ref} = (\mathbf{V}_{ref}, \mathbf{E}_{ref})$; 
            The number of simulation steps for this moving part in the current convex $N_s$. 
            Joint information set $\mathcal{J}$ of the moving part $p_{mov}$. 
        \Ensure
            Average penetration depth APD ($\mathcal{L}_{phy}^{cur}$) for this part $p_{mov}$ in the current context. 
            Projection loss $\mathcal{L}_{proj}^{cur}$ for this part $p_{mov}$ in the current context. 
        \If{$\mathcal{J}$.moving\_type == NONE\_MOTION}  \\ 
            \Return 0 , 0
        \EndIf
        \State $\mathbf{N}_{ref} \leftarrow \text{Face-Normal}(\mathbf{V}_{ref}, \mathbf{E}_{ref})$
        \State $\mathbf{d}^0 \leftarrow \text{Vertex-Face-Distance}(\mathbf{V}_{mov}, \mathbf{V}_{ref}, \mathbf{E}_{ref}, \mathbf{N}_{ref})$ 
        \State $\mathbf{S}^0 \leftarrow \text{Sign}(\mathbf{d}^0)$
        \State $\mathcal{S}_{phy}\leftarrow \emptyset$
        \State $\mathcal{S}_{proj}\leftarrow \emptyset$
        \State Let $[l, u]$ be the articulation range of $p_{mov}$ which is contained in the joint information $\mathcal{J}$
        \For{$t = 1$ to $N_{s}$}
            \State $s^t\leftarrow l + (u - l) \cdot \frac{t}{N_s}$
            \State $p_{mov}^t = (\mathbf{V}_{mov}^t, \mathbf{E}_{mov})\leftarrow \text{Articulation-Simulation}(\mathbf{V}_{mov}, s^t, \mathcal{J})$ 
            \State $\mathbf{D}^t \leftarrow \text{Vertex-Face-Distance}(\mathbf{V}_{mov}^t, \mathbf{V}_{ref}, \mathbf{E}_{ref}, \mathbf{N}_{ref})$
            \State $\mathbf{S}^t \leftarrow \text{Sign}(\mathbf{D}^t)$
            \State $\mathbf{C}^t\leftarrow \text{Vertices-In-Faces}(\mathbf{V}_{mov}^t, \mathbf{V}_{ref}, \mathbf{E}_{ref})$
            \State $\mathbf{C}^t \leftarrow \mathbf{C}^t \wedge (\mathbf{S}^t \neq  \mathbf{S}^{t-1} )$
            \State $\text{PeneD}(p_{mov}^t, p_{ref}) \leftarrow \text{Mean}(\mathbf{C}^t \cdot \mathbf{D}^t \cdot \mathbf{S}^t, \text{dim}=0,1 )$ 
            \State $\mathcal{S}_{phy} \leftarrow \mathcal{S}_{phy}  \cup \{  \text{PeneD}(p_{mov}^t, p_{ref}) \}$

            \State $\Delta \mathbf{V}_{mov}\leftarrow \mathbf{V}_{mov}^t - \mathbf{V}_{mov}^{t-1}$
            \State $\text{ProjD} (p_{mov}^t, p_{ref}) \leftarrow \text{Mean}(\text{Sum}( \text{Expand}( \Delta \mathbf{V}_{mov}, \text{dim}=1) \cdot \text{Expand}( \mathbf{C}^t \cdot \mathbf{D}^t, \text{dim=2}) \cdot \text{Expand} (\mathbf{N}_{ref}, \text{dim}=0 ), \text{dim} = 2), \text{dim}=0,1)$
            \State $\mathcal{S}_{proj} \leftarrow  \mathcal{S}_{proj}  \cup \{  \text{ProjD} (p_{mov}^t, p_{ref})  \}$
        \EndFor\\
        \State $\mathcal{L}_{phy}^{cur} \leftarrow \text{Average}(\mathcal{S}_{phy})$
        \State $\mathcal{L}_{proj}^{cur} \leftarrow \text{Average}(\mathcal{S}_{proj})$ \\ 
        \Return $\mathcal{L}_{phy}^{cur}$ , $\mathcal{L}_{proj}^{cur}$
    \end{algorithmic}
\end{algorithm}

\begin{algorithm}[H]
\small 
\caption{\textbf{Physics-aware losses}.
}
\label{algo_phy_sim}
\footnotesize
    \begin{algorithmic}[1]
        \Require
            An articulated mesh $a$ with a set of moving parts $\{  p_1, ..., p_{k_p} \}$ and their joint information $\{ \mathcal{J}_1, ..., \mathcal{J}_{k_p} \}$, where $k_p$ is the number of parts in $a$. 
            Number of simulation steps $N_{s}$ for a single part articulation simulation process, \emph{i.e.,} moving one part when other parts are put into a specific articulated state. 
            The number of single-part articulation simulation processes $N_{det}$ for each part. 
        \Ensure
            Average penetration depth APD ($\mathcal{L}_{phy}$). 
            Projection loss $\mathcal{L}_{proj}$. 
        \State $\mathcal{S}_{phy}\leftarrow \emptyset$
        \State $\mathcal{S}_{proj}\leftarrow \emptyset$
        \For{$p_{mov} =   p_1$ to $p_{k_p}$}
            \For{$i_{step} = 1$ to $N_{det}$}
                \State Sample an articulation state for each part expect for $p_{mov}$: $\{ s_{p}  \vert p \in \{  p_1, ..., p_{k_p} \}, p \neq p_{mov} \}$
                \State Put parts except for $p_{mov}$ into their sampled states; Put $p_{mov}$ into its rest articulation state. 
                \State $p_{ref} \leftarrow \text{Merge-Meshes}(\{  p_1, ..., p_{k_p} \} \setminus \{ p_{mov}\} )$
                \State  
                $\mathcal{L}_{phy}^{cur}, \mathcal{L}_{proj}^{cur} \leftarrow \text{Single-Simulation}(p_{mov}, p_{ref}, N_s, \mathcal{J}_{p_{mov}})$
                \State $\mathcal{S}_{phy} \leftarrow  \mathcal{S}_{phy} \cup \{  \mathcal{L}_{phy}^{cur}\}$
                \State $\mathcal{S}_{proj} \leftarrow  \mathcal{S}_{proj}  \cup \{  \mathcal{L}_{proj}^{cur} \}$
            \EndFor
        \EndFor\\
        \State $\mathcal{L}_{phy} \leftarrow \text{Average}(\mathcal{S}_{phy})$
        \State $\mathcal{L}_{proj} \leftarrow \text{Average}(\mathcal{S}_{proj})$ \\ 
        \Return $\mathcal{L}_{phy}$ , $\mathcal{L}_{proj}$
        
    \end{algorithmic}
\end{algorithm}

\vpara{How to use $\mathcal{L}_{proj}$ to optimize the articulated mesh $a$?} 
Since the calculation process of $\mathcal{L}_{proj}$ is differentiable, we can update shape $a$ by back-propagating $\mathcal{L}_{proj}$ to update the global shape deformation parameter $\mathbf{z}$. 
Specifically, we calculate the gradient of $\mathcal{L}_{proj}$ over $\mathbf{z}$, which can be easily realized by the support of PyTorch's ``autograd'' package, and then update $\mathbf{z}$ via the gradient, \emph{i.e.,} $\mathbf{z}\leftarrow \mathbf{z} - \epsilon_{proj} \frac{\partial \mathcal{L}_{proj}}{\partial \mathbf{z}}$. 
$\epsilon_{proj}$ serves as the ``learning rate'' for the deformation coefficient $\mathbf{z}$ here and is set to $10^{-4}$ at the training time and $10^{-5}$ at the inference/sampling time in our implementation.

\subsection{Additional Explanations} \label{sec_supp_method_additional_details}

\begin{figure*}[ht]
    \centering
      \includegraphics[width=1.0\textwidth]{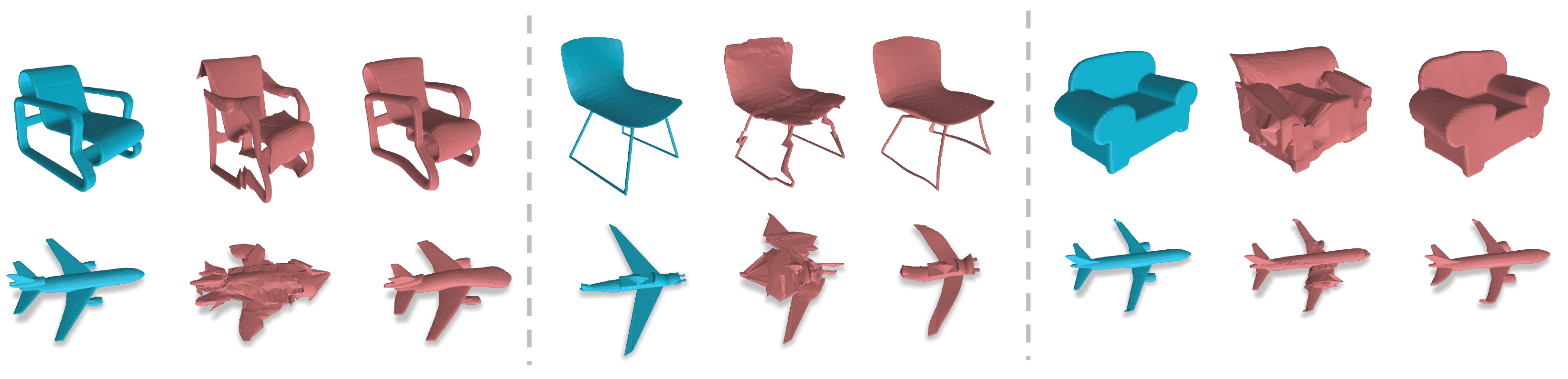}
    \caption{ \footnotesize
        \textbf{The effectiveness of the mesh smooth layer.} 
        For every three shapes, the \textbf{leftmost} one is the reference shape for deformation, the \textbf{middle} one is the generated mesh without smoothing, and the \textbf{right} one is the shape after smoothing. 
        }
    \label{fig_smooth_layer}
\end{figure*}


\vpara{Mesh smooth layer.}
Directly composing deformed convexes together for object-level meshes would usually lead to unwanted artifacts near convex edges, as shown in each middle one of every three shapes in Figure~\ref{fig_smooth_layer}.  
To tackle this issue and to produce smooth object-level meshes, we add a smooth layer following~\cite{garcia2013cages}, resulting in the rightmost shape in every three shapes drawn in Figure~\ref{fig_smooth_layer}.

\section{Additional Experiments} \label{sec_supp_exp}


\subsection{Few-Shot Mesh Generation} \label{sec_supp_few_shot_mesh_gen}

We further discuss our few-shot mesh generation performance from the following aspects to validate the merits of our hierarchical mesh deformation-based generative scheme: 
\begin{itemize}
    \item Few-shot generation performance w.r.t. the number of observed reference examples (\#Shots);
    \item Few-shot generation performance on rigid categories;
    \item Additional results on human bodies. 
\end{itemize}

\begin{table*}[t]
    \centering
    \caption{\footnotesize 
    \textbf{Experimental comparisons on Eyeglasses category.}
    MMD is multiplied by $10^3$. 
    \textbf{Bold} numbers for best values. 
    } 
    \begin{tabular}{@{\;}c@{\;}|c|c|c|c|c@{\;}}
    \midrule
        \hline
        \specialrule{0em}{1pt}{0pt} 
        \#Shots & Method & MMD~($\downarrow$) & COV~(\%, $\uparrow$) &  1-NNA~(\%, $\downarrow$)  & JSD~($\downarrow$)  \\ 
        \cline{1-6} 
        \specialrule{0em}{1pt}{0pt}
        
        \multirow{3}{*}{2}  & PolyGen~\cite{nash2020polygen} & 9.558 & 3.18 & 99.70 & 0.1543
        \\ \cline{2-6} 
        \specialrule{0em}{1pt}{0pt}

        ~ & DeepMetaHandles~\cite{liu2021deepmetahandles} &  8.684 & 6.67 & 99.21 & 0.1585
        \\ \cline{2-6} 
        \specialrule{0em}{1pt}{0pt}

        ~ & Ours & \textbf{7.279} & \textbf{19.30} & \textbf{98.41} & \textbf{0.0903}  
        \\ \cline{1-6} 
        \specialrule{0em}{1pt}{0pt}

        \multirow{3}{*}{4}   & PolyGen~\cite{nash2020polygen} & 8.669 & 8.28 & 98.42 & 0.1425
        \\ \cline{2-6} 
        \specialrule{0em}{1pt}{0pt}

        ~ & DeepMetaHandles~\cite{liu2021deepmetahandles} &  6.685 & 12.57 & \textbf{98.50} & 0.1036
        \\ \cline{2-6} 
        \specialrule{0em}{1pt}{0pt}

        ~ & Ours & \textbf{6.303} & \textbf{27.54} & {98.80} & \textbf{0.0840} 
        \\ \cline{1-6} 
        \specialrule{0em}{1pt}{0pt}

        \multirow{3}{*}{8}   & PolyGen~\cite{nash2020polygen} & 7.663 & 15.56 & 97.24 & 0.1054
        \\ \cline{2-6} 
        \specialrule{0em}{1pt}{0pt}

        ~ & DeepMetaHandles~\cite{liu2021deepmetahandles} &  6.222 &17.33 &97.99 & 0.0864
        \\ \cline{2-6} 
        \specialrule{0em}{1pt}{0pt}

        ~ & Ours & \textbf{6.102} & \textbf{34.56} & \textbf{97.25} & \textbf{0.0799}
        \\ \cline{1-6} 
        \specialrule{0em}{1pt}{0pt}

    \end{tabular}
    \label{tb_exp_fs_gen_rigid_eyeglasses}
\end{table*}

\begin{table*}[t]
    \centering
    \caption{\footnotesize 
    \textbf{Experimental comparisons on Scissors category.}
    MMD is multiplied by $10^3$. 
    \textbf{Bold} numbers for best values. 
    } 
    \begin{tabular}{@{\;}c@{\;}|c|c|c|c|c@{\;}}
    \midrule
        \hline
        \specialrule{0em}{1pt}{0pt} 
        \#Shots & Method & MMD~($\downarrow$) & COV~(\%, $\uparrow$) &  1-NNA~(\%, $\downarrow$)  & JSD~($\downarrow$)  \\ 
        \cline{1-6} 
        \specialrule{0em}{1pt}{0pt}
        
        \multirow{3}{*}{2}  & PolyGen~\cite{nash2020polygen} & 7.311 & 4.55 & 99.65 & 0.5192
        \\ \cline{2-6} 
        \specialrule{0em}{1pt}{0pt}

        ~ & DeepMetaHandles~\cite{liu2021deepmetahandles} &  6.154 & 12.19 & 98.39 & 0.3315
        \\ \cline{2-6} 
        \specialrule{0em}{1pt}{0pt}

        ~ & Ours & \textbf{2.503} & \textbf{26.19} & \textbf{98.31} & \textbf{0.1412}  
        \\ \cline{1-6} 
        \specialrule{0em}{1pt}{0pt}

        \multirow{3}{*}{4}   & PolyGen~\cite{nash2020polygen} & 4.015 & 9.52 & 98.96 & 0.3459
        \\ \cline{2-6} 
        \specialrule{0em}{1pt}{0pt}

        ~ & DeepMetaHandles~\cite{liu2021deepmetahandles} &  1.875 & 24.20 & 98.76 & 0.2173
        \\ \cline{2-6} 
        \specialrule{0em}{1pt}{0pt}

        ~ & Ours & \textbf{1.534} & \textbf{50.45} & \textbf{97.81} & \textbf{0.1299} 
        \\ \cline{1-6} 
        \specialrule{0em}{1pt}{0pt}

        \multirow{3}{*}{8}   & PolyGen~\cite{nash2020polygen} &3.108 & 13.16 & 97.91 & 0.2067
        \\ \cline{2-6} 
        \specialrule{0em}{1pt}{0pt}

        ~ & DeepMetaHandles~\cite{liu2021deepmetahandles} &  1.747 & 32.19 & 96.76 & 0.2017
        \\ \cline{2-6} 
        \specialrule{0em}{1pt}{0pt}

        ~ & Ours & \textbf{1.184} & \textbf{63.47} & \textbf{96.12} & \textbf{0.1256}
        \\ \cline{1-6} 
        \specialrule{0em}{1pt}{0pt}

    \end{tabular}
    \label{tb_exp_fs_gen_rigid_scissors}
\end{table*}

\begin{table*}[t]
    \centering
    \caption{\footnotesize 
    \textbf{Experimental comparisons on TrashCan category.}
    MMD is multiplied by $10^3$. 
    \textbf{Bold} numbers for best values. 
    } 
    \begin{tabular}{@{\;}c@{\;}|c|c|c|c|c@{\;}}
    \midrule
        \hline
        \specialrule{0em}{1pt}{0pt} 
        \#Shots & Method & MMD~($\downarrow$) & COV~(\%, $\uparrow$) &  1-NNA~(\%, $\downarrow$)  & JSD~($\downarrow$)  \\ 
        \cline{1-6} 
        \specialrule{0em}{1pt}{0pt}
        
        \multirow{3}{*}{2}  & PolyGen~\cite{nash2020polygen} & 16.448 & 4.29 & 97.44 & 0.3224
        \\ \cline{2-6} 
        \specialrule{0em}{1pt}{0pt}

        ~ & DeepMetaHandles~\cite{liu2021deepmetahandles} &  14.589 & 6.33 & 90.24 & 0.2170
        \\ \cline{2-6} 
        \specialrule{0em}{1pt}{0pt}

        ~ & Ours & \textbf{12.746} & \textbf{9.00} & \textbf{84.81} & \textbf{0.1436}  
        \\ \cline{1-6} 
        \specialrule{0em}{1pt}{0pt}

        \multirow{3}{*}{4}   & PolyGen~\cite{nash2020polygen} & 10.218 & 13.22 & 93.09 & 0.2331
        \\ \cline{2-6} 
        \specialrule{0em}{1pt}{0pt}

        ~ & DeepMetaHandles~\cite{liu2021deepmetahandles} &  9.402 & \textbf{17.14} & 86.49 & 0.1913
        \\ \cline{2-6} 
        \specialrule{0em}{1pt}{0pt}

        ~ & Ours & \textbf{8.499} & \textbf{17.14} & \textbf{73.11} & \textbf{0.1003}
        \\ \cline{1-6} 
        \specialrule{0em}{1pt}{0pt}

        \multirow{3}{*}{8}   & PolyGen~\cite{nash2020polygen} & 9.001 & 16.67 & 90.79 & 0.1938
        \\ \cline{2-6} 
        \specialrule{0em}{1pt}{0pt}

        ~ & DeepMetaHandles~\cite{liu2021deepmetahandles} &  8.970 & 22.83 & 84.44 & 0.1379
        \\ \cline{2-6} 
        \specialrule{0em}{1pt}{0pt}

        ~ & Ours & \textbf{8.134} & \textbf{24.47} & \textbf{71.02} & \textbf{0.0935}
        \\ \cline{1-6} 
        \specialrule{0em}{1pt}{0pt}

    \end{tabular}
    \label{tb_exp_fs_gen_rigid_trashcan}
\end{table*}

\vpara{Few-shot generation performance w.r.t. \#Shots.}
For Scissors, Eyeglasses, and TrashCan containing relatively rich objects, we try to vary the value of the number of observed examples (\#Shots) and compare the performance achieved by different methods on each few-shot setting. 
We consider three additional settings, namely 2-shots, 4-shots, and 8-shots. 
To make results comparable across different few-shot settings, we use the \textbf{same test set} for all of those settings. 
From Table~\ref{tb_exp_fs_gen_rigid_eyeglasses}, ~\ref{tb_exp_fs_gen_rigid_scissors}, ~\ref{tb_exp_fs_gen_rigid_trashcan}, 
we can observe that our method can consistently outperform baseline strategies. 
It successfully guides the model to generate diverse samples even under the 2-shots setting, bypassing baselines by a large margin, \emph{e.g.,} 189\% relatively higher coverage ratio than DeepMetaHandles on the Eyeglasses category. 
Notice that it is not at a cost of sacrificing the mesh quality, \emph{i.e.,} we can achieve the lowest MMD scores at the same time on all of those three categories. 

Further, as a general empirical rule observed from the results, 
more observed examples would always lead to better few-shot generation performance, guiding the model to generate visually plausible samples with higher diversity. 
It aligns well with our intuitions and also conclusions made in few-shot image generation works~\cite{hong2022deltagan,gu2021lofgan}. 

\begin{figure*}[ht]
    \centering
      \includegraphics[width=1.0\textwidth]{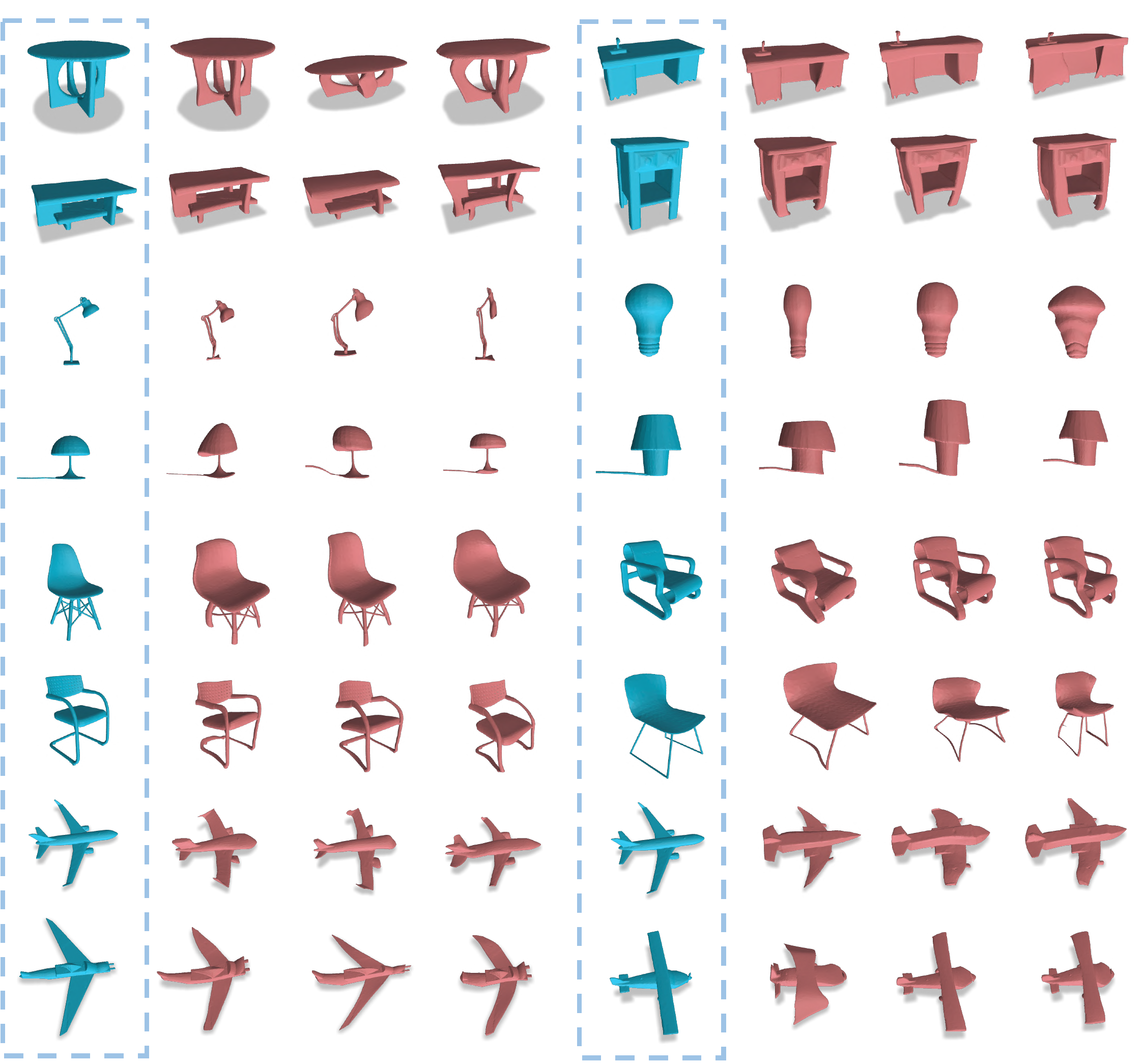}
    \caption{ \footnotesize
        \textbf{Qualitative evaluation on few-shot rigid mesh generation.}
        For every four shapes, the leftmost one (highlighted by \textcolor{myblue}{blue rectangles} ) is the reference shape from the training set, while the remaining three are conditionally generated samples. Object categories from top to down are Table, Lamp, Chair, and Airplane. 
        }
    \label{fig_free_gen_vis_rigid}
\end{figure*}

\begin{table*}[t]
    \centering
    \caption{\footnotesize 
     \textbf{Experimental comparisons on rigid mesh categories.}
    MMD is multiplied by $10^3$. 
    \textbf{Bold} numbers for best values.
    } 
    \begin{tabular}{@{\;}c@{\;}|c|c|c|c|c|c@{\;}}
    \midrule
        \hline
        \specialrule{0em}{1pt}{0pt} 
        Category & \#Shots & Method & MMD~($\downarrow$) & COV~(\%, $\uparrow$) &  1-NNA~(\%, $\downarrow$)  & JSD~($\downarrow$)  \\ 
        \cline{1-7} 
        \specialrule{0em}{1pt}{0pt}

        \multirow{4}{*}{Table} &  2 & Ours & 6.162 & 8.20 & 98.87 & 0.1118
        \\ \cline{2-7} 
        \specialrule{0em}{1pt}{0pt}

        ~ & 4 & Ours & 4.206 & 13.63 & 97.80 & 0.0763 
        \\ \cline{2-7} 
        \specialrule{0em}{1pt}{0pt}

        ~ & \multirow{2}{*}{8}  & DeepMetaHandles~\cite{liu2021deepmetahandles} &  6.640 & 7.45 & 99.34 & 0.1046
        \\ \cline{3-7} 
        \specialrule{0em}{1pt}{0pt}

        ~ & ~ & Ours & \textbf{2.356} & \textbf{25.23} &\textbf{96.33} & \textbf{0.0479}
        \\ \cline{1-7} 
        \specialrule{0em}{1pt}{0pt}

        \multirow{4}{*}{Chair} &  2 & Ours & 4.148 & 15.20 & 99.42 & 0.1080  
        \\ \cline{2-7} 
        \specialrule{0em}{1pt}{0pt}

        ~ & 4 & Ours & 3.363 & 19.36 & 98.63 & 0.0512 
        \\ \cline{2-7} 
        \specialrule{0em}{1pt}{0pt}

        ~ &  \multirow{2}{*}{8} & DeepMetaHandles~\cite{liu2021deepmetahandles} &  5.205 & 12.70 & 99.43 & 0.0738
        \\ \cline{3-7} 
        \specialrule{0em}{1pt}{0pt}

        ~ &  ~ & Ours & \textbf{2.690} & \textbf{28.43} & \textbf{97.09} & \textbf{0.0294}
        \\ \cline{1-7} 
        \specialrule{0em}{1pt}{0pt}

        \multirow{4}{*}{Lamp} & 2 & Ours & 4.029 & 16.80 & 97.17 & 0.0930
        \\ \cline{2-7} 
        \specialrule{0em}{1pt}{0pt}

        ~ & 4 & Ours & 3.556 & 20.70 & 96.13 & 0.0749 
        \\ \cline{2-7} 
        \specialrule{0em}{1pt}{0pt}

         ~ & \multirow{2}{*}{8}  & DeepMetaHandles~\cite{liu2021deepmetahandles} &  9.730 & 11.40 & 99.75 & 0.1966
        \\ \cline{3-7} 
        \specialrule{0em}{1pt}{0pt}

        ~ & ~ & Ours & \textbf{2.822} & \textbf{30.20} & \textbf{93.67} & \textbf{0.0671}
        \\ \cline{1-7} 
        \specialrule{0em}{1pt}{0pt}

        \multirow{4}{*}{Airplane} &  2 & Ours & 1.029 & 15.93 & 97.47 & 0.0906
        \\ \cline{2-7} 
        \specialrule{0em}{1pt}{0pt}

        ~ & 4 & Ours & 0.927 & 21.15 & 97.33 & 0.0401 
        \\ \cline{2-7} 
        \specialrule{0em}{1pt}{0pt}

        ~ & \multirow{2}{*}{8}  & DeepMetaHandles~\cite{liu2021deepmetahandles} &  2.295 & 9.37 & 99.31 & 0.1654
        \\ \cline{3-7} 
        \specialrule{0em}{1pt}{0pt}

        ~ & ~ & Ours & \textbf{0.869} & \textbf{25.71 }& \textbf{97.08} & \textbf{0.0334}
        \\ \cline{1-7} 
        \specialrule{0em}{1pt}{0pt}

    \end{tabular}
    \label{tb_exp_fs_gen_rigid_all}
\end{table*}

\vpara{Few-shot generation performance on rigid mesh categories.} 
To further prove the effectiveness of our hierarchical mesh deformation-based generative strategy as a general few-shot generation method not restricted to articulated objects, 
we test it on four rigid categories from the ShapeNet dataset~\cite{chang2015shapenet} (\emph{i.e.}, Table, Chair, Lamp, and Airplane) and compare it to the baselines. 
Categories for pre-training convex-level deformation models of  Table, Chair,  Lamp, and Airplane are Chair, Table, Airplane, and Lamp respectively. 
From 
Table~\ref{tb_exp_fs_gen_rigid_all}, 
our method can still achieve better performance on all of those four categories. 
It further demonstrates the superiority of our hierarchical mesh deformation strategy as a general few-shot generation method over previous methods. 
We also report the results achieved by our method on the other two few-shot learning settings. 
We can make similar observations by comparing across different few-shot learning settings. 
As a qualitative evaluation, we draw samples from our model for the above four categories in Figure~\ref{fig_free_gen_vis_rigid} (under the 4-shot setting). 

\begin{figure}[ht]
    \centering
      \includegraphics[width=0.45\textwidth]{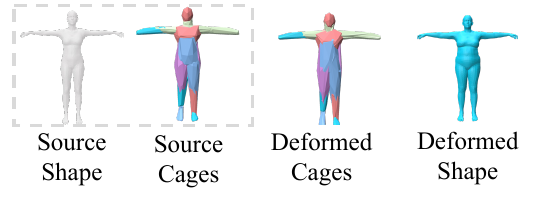}
    \caption{ \footnotesize
        \textbf{Additional deformation results on human bodies.}
        }
    \label{fig_deform_human_bodies}
\end{figure}

\vpara{Deformation results on human bodies}
Ours approach  can indeed generalize to complex shape such as human bodies as exampled in Figure~\ref{fig_deform_human_bodies}. We mainly leverage this example to show the method is not restricted to relatively simple piece-wise rigid objects demonstrated in main experiments. However, we do not conduct abundant experiments on it since human bodies are deformable and have rich data sources, which diverge from our focus on piece-wise rigid objects with limited examples. 

\subsection{Transfer Learning and Fine-tuning for Convex Deformations} \label{sec_supp_tranfer_cvx_def}

\begin{table*}[htbp]
    \centering
    \caption{\footnotesize 
    \textbf{Few-shot generation performance w.r.t. source categories for transfer learning} on Eyeglasses, Scissors, and TrashCan categories. ``All'' denotes using all of those four categories as the source category. 
    } 
    \begin{tabular}{@{\;}c@{\;}|c|c|c|c|c@{\;}}
    \midrule
        \hline
        \specialrule{0em}{1pt}{0pt} 
        \makecell[c]{Target \\ Category} & \makecell[c]{Source \\ Category} & MMD~($\downarrow$) &  COV~(\%, $\uparrow$)  & 1-NNA~(\%, $\downarrow$) & JSD~($\downarrow$)\\ 
        \cline{1-6} 
        \specialrule{0em}{1pt}{0pt}

        \multirow{5}{*}{Eyeglasses} & Table & 6.07 & 25.00 & 99.13 & 0.0875 
        \\ \cline{2-6} 
        \specialrule{0em}{1pt}{0pt}

        ~ & Chair & 8.23 & 20.53 & 99.57 & 0.1188 
        \\ \cline{2-6} 
        \specialrule{0em}{1pt}{0pt}

        ~ & Lamp & 6.47 & 26.67 & 98.46 & 0.0973 
        \\ \cline{2-6} 
        \specialrule{0em}{1pt}{0pt}

        ~ & Airplane & 8.28 & 21.91 & 99.35 & 0.1042 
        \\ \cline{2-6} 
        \specialrule{0em}{1pt}{0pt}

        ~ &  All & \textbf{6.06} & \textbf{29.82} & \textbf{98.26} & \textbf{0.0681} 
        \\ \cline{1-6} 
        \specialrule{0em}{1pt}{0pt}

        \multirow{5}{*}{Scissors} & Table & 1.63 & 51.07 & 97.55 & 0.1579 
        \\ \cline{2-6} 
        \specialrule{0em}{1pt}{0pt}

        ~ & Chair & 1.57 & 56.76 & 97.10 & 0.1752 
        \\ \cline{2-6} 
        \specialrule{0em}{1pt}{0pt}

        ~ & Lamp & \textbf{1.34} & 54.39 & 97.79 & 0.1328 
        \\ \cline{2-6} 
        \specialrule{0em}{1pt}{0pt}

        ~ & Airplane & 1.67 & 52.19 & 97.36 & 0.1724 
        \\ \cline{2-6} 
        \specialrule{0em}{1pt}{0pt}

        ~ &  All & {1.50} & \textbf{57.89} & \textbf{97.02} & \textbf{0.1274} 
        \\ \cline{1-6} 
        \specialrule{0em}{1pt}{0pt}

        \multirow{5}{*}{TrashCan} & Table & 8.43 & 17.07 & 72.87 & \textbf{0.0933} 
        \\ \cline{2-6} 
        \specialrule{0em}{1pt}{0pt}

        ~ & Chair & \textbf{8.39} & 17.07 & 74.29 & 0.0939 
        \\ \cline{2-6} 
        \specialrule{0em}{1pt}{0pt}

        ~ & Lamp & 9.49 & \textbf{19.29} & 73.58 & 0.1138 
        \\ \cline{2-6} 
        \specialrule{0em}{1pt}{0pt}

        ~ & Airplane & 13.20 & 10.32 & 74.29 & 0.1429 
        \\ \cline{2-6} 
        \specialrule{0em}{1pt}{0pt}

        ~ &  All & {8.43} & {17.14} & \textbf{72.09} & {0.0994} 
        \\ \cline{1-6} 
        \specialrule{0em}{1pt}{0pt}

    \end{tabular}
    \vspace{-4pt}
    \label{tb_exp_transfer_all}
\end{table*} 

We examine the role of transfer learning and fine-tuning in our method and find that 1)  \textbf{Transfer learning}'s power can be boosted by increasing the amount of source data and is affected by the affinity between source and target categories; 2) \textbf{Fine-tuning} can help learn category-specific deformation patterns, benefiting quality and diversity. 



\vpara{Transfer learning w.r.t. amount of data for transferring.}
Table~\ref{tb_exp_fine_tuning} compares the model's performance when a) using all source data for transferring, b) using half amount of total source data, and c) use no transfer learning. The transfer learning's effectiveness can be boosted increasing the amount of source data for transferring. 

\vpara{Transfer learning w.r.t. source categories.}
To test the influence of source categories for deformation pattern transferring on the few-shot generation performance, we try to vary the source categories by using each category individually as the source and comparing them. 
We conduct experiments on three relatively rich articulated mesh categories, namely Eyeglasses, Scissors, and TrashCan. 
From Table~\ref{tb_exp_transfer_all}, we can make the following observations: 
1) For Scissors, Lamp is a friendly category to reduce the minimum matching distance for samples of higher quality. 
2) For TrashCan, Lamp as the source category can help with enhancing the diversity of generated samples, perhaps due to diverse deformation patterns transferred to the TrashCan's body parts. 

\vpara{Fine-tuning.}
The effectiveness of the fine-tuning process can be examined by comparing the ablated version and the full model demonstrated in Table~\ref{tb_exp_fine_tuning}.  

\begin{table*}[htbp]
    \centering
    \caption{\footnotesize  Ablations w.r.t. the effectiveness of pre-training and fine-tuning. 
    }
        \begin{tabular}{@{\;}l@{\;}|c|c|c|c|c@{\;}}
        \toprule
        Method & MMD~($\downarrow$) &  COV~(\%, $\uparrow$)  & 1-NNA~(\%, $\downarrow$) & JSD~($\downarrow$) & APD~($\downarrow$) \\ 
        \midrule

        Ours w/o Transfer & 5.424 & 46.64 & 93.01 & 0.1159 & {1.3822}
        \\ 
        \midrule

        Ours w/ Transfer (Half Data) & 5.201 & 49.43 & 92.81 & 0.1130 & {1.3365}
        \\ 
        \midrule

        Ours w/o Fine-tuning & 6.538 & 43.20 & 94.70 & 0.1437 & {1.4530}
        \\ 
        \midrule
        
        Ours & \textbf{5.198} & \textbf{50.45} & \textbf{92.36} & \textbf{0.1118} & \textbf{1.3192}
        \\
        \bottomrule
        
    \end{tabular}
    \label{tb_exp_fine_tuning}
\end{table*}

\begin{table*}[htbp]
    \centering
    \caption{\footnotesize 
    \textbf{Comparison between methods using the transfer learning strategy at the object level and the convex level. }
    For metrics of each version, we report their average value over all categories. 
    MMD is multiplied by $10^3$ and APD is multiplied by $10^2$. 
    \textbf{Bold} numbers for best values.  
    } 
    \begin{tabular}{@{\;}c@{\;}|c|c|c|c@{\;}}
    \midrule
        \hline
        \specialrule{0em}{1pt}{0pt} 
        Method & MMD~($\downarrow$) &  COV~(\%, $\uparrow$)  & 1-NNA~(\%, $\downarrow$) & JSD~($\downarrow$)\\ 
        \cline{1-5} 
        \specialrule{0em}{1pt}{0pt}

        Ours (Transfer Obj.) & 7.339 & 32.70 & 96.11 & 0.1557 
        \\ \cline{1-5} 
        \specialrule{0em}{1pt}{0pt}

        Ours w/o Hier. & 7.170 & 36.41 & 95.43 & 0.1492 
        \\ \cline{1-5} 
        \specialrule{0em}{1pt}{0pt}

         Ours & \textbf{5.198} & \textbf{50.45} & \textbf{92.36} & \textbf{0.1118} 
        \\ \cline{1-5} 
        \specialrule{0em}{1pt}{0pt}

    \end{tabular}
    \vspace{-4pt}
    \label{tb_exp_gen_comparison_trnaferring}
\end{table*}

\begin{figure}[ht]
    \centering
    \includegraphics[width=0.28\textwidth]{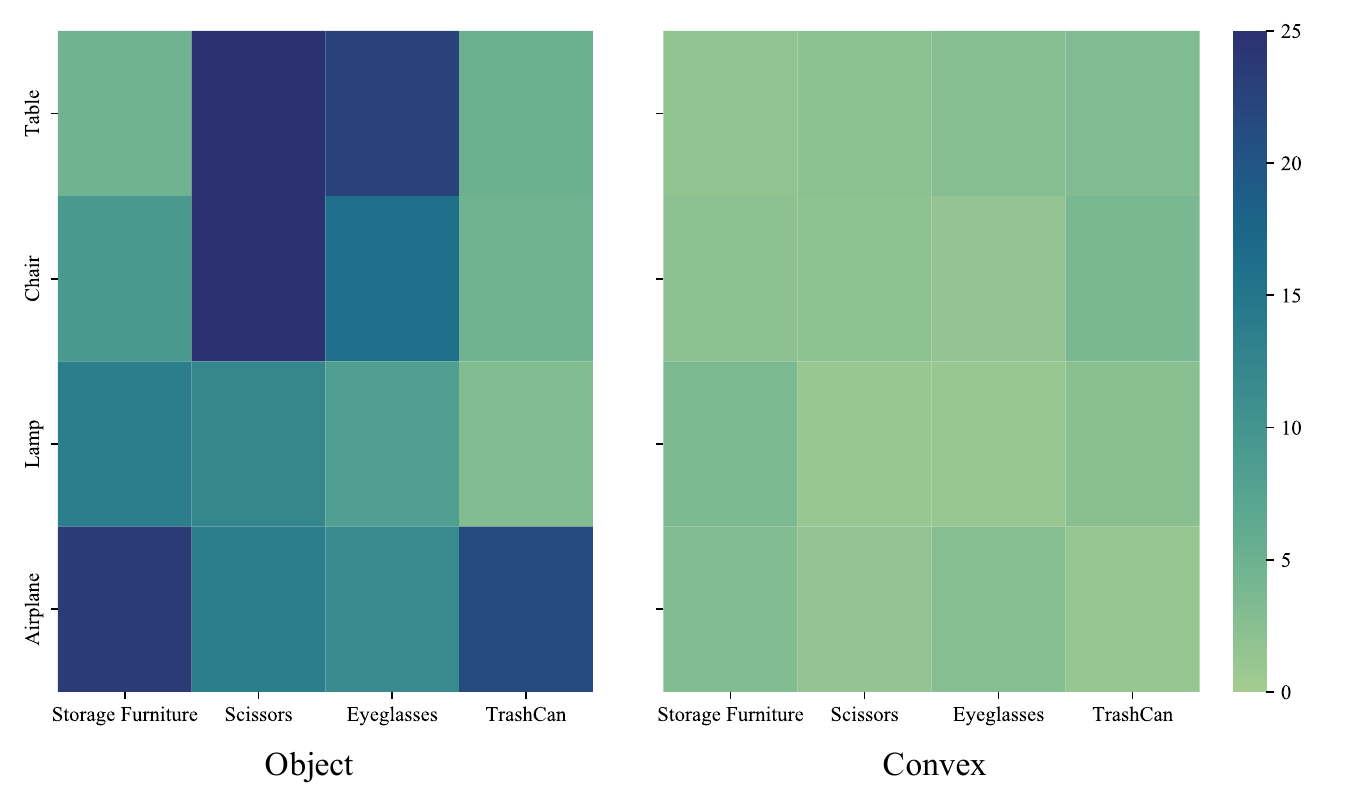}
    \vspace{-5pt}
    \caption{ \footnotesize
        \textbf{Domain gap measured by different levels of shapes.}
        Heatmaps of minimum matching distances between pre-training and target datasets. 
        }
        \vspace{-20pt}
    \label{fig_mmd_dist_shp_cvx}
\end{figure}

\vpara{Intermediates for transferring deformation patterns.}
We choose to use convexes as intermediates to learn and transfer shared deformation patterns in this work. 
It comes from the assumption and the observation that the convex distribution is more similar across different categories than the whole shape distribution (as shown in Figure~\ref{fig_mmd_dist_shp_cvx}). 
Therefore, from our intuition, convexes are more promising to serve as intermediates for transferring mesh deformation patterns across different categories. 
We further validate this intuition by trying to learn and transfer deformation patterns at the object level. 
As shown in Table~\ref{tb_exp_gen_comparison_trnaferring}, our model using convexes as intermediates can perform much better than the trail on transferring at the object level. 
This validates one of our crucial assumptions in this work.

Besides, we cannot even observe trivially transferring deformation patterns at the object level as a better strategy than the method without any deformation transferring. 
Usually, transferring deformation rules for the whole object requires some special designs~\cite{sung2020deformsyncnet}. 
This confirms the value of our hierarchical deformation design for mesh deformation transferring.

\begin{figure}[ht]
    \centering
      \includegraphics[width=0.45\textwidth]{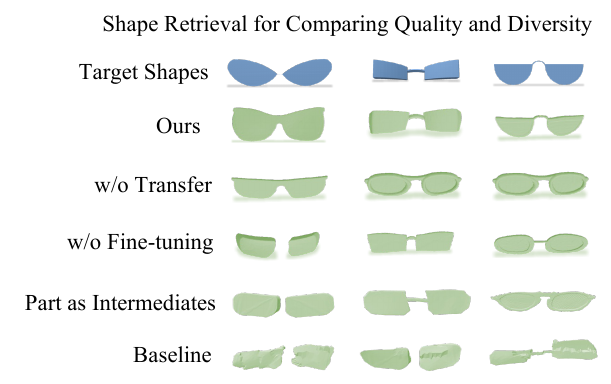}
    \caption{ \footnotesize
        \textbf{Shape retrieval for comparing quality and diversity.}
        }
    \label{fig_shape_retreieval_exp}
\end{figure}

\vpara{Shape retrieval experiments for comparing quality and diversity.}
As a further intuitive demonstration on the effectiveness of our techniques (\emph{i.e.,} transfer learning, fine-tuning, and convex as intermediates ) on improving the results' diversity and quality transfer learning, we conduct a shape retrieval experiment and compare our model with different ablated versions. 
Given each target shape, we select the its closest shape from generated assets as the retrieval results. 
The results are presented in Figure~\ref{fig_shape_retreieval_exp}. Only ours can give results that are plausible, closet to the target, while also different from each other. 


\subsection{Generation via Deformation} \label{sec_supp_generation_via_deformation}

There are a variety of generation techniques that have been developed recently in the tide of AIGC, such as score-based generative models and diffusion models. 
Along with them, many works try to explore the possibility of leveraging such techniques for 3D content generation. 
They mostly aim at generating shapes as a whole without considering their functionalities. 

In this work, we wish to generate physically-realistic articulated meshes and resort to a relatively traditional generation technique, \emph{i.e.,} generation via deformation. 
The reasons of our choice are mainly as follows: 
\textbf{1)} Deformation-based generation could let us parameterize shape variations into a low-dimensional space, by leveraging cages to control deformations and by using deformation bases to record common deformation patterns. 
\textbf{2)} Deformation-based generation could easily let us find suitable intermediates such that shared deformation patterns can be transferred across different categories easily. 
\textbf{3)} It also enables us to devise an effective physics-aware correction scheme to guide the model to generate physically-realistic deformations. 

To explore more possibilities, we try to design another method that leverages point cloud diffusion, surface reconstruction, and a physics-aware correction designed for the reverse diffusion process, supported by the differentiable surface reconstruction algorithm. 
However, sometimes we observe that it tends to produce meshes of poor quality. 
We suppose that it comes from the difficulty to impose physics-related constraints on the generated meshes from point cloud diffusion. 

We would elaborate on details of this method and its results in the following text. 

\vpara{Method: Hierarchical generation via point cloud diffusion.} 
It proceeds as follows: 
1) Represent the shape via an object-convex hierarchy via convex decomposition. 
2) Train a convex-level conditional point cloud diffusion model on source categories.
3) Transfer the pre-trained convex-level point cloud diffusion model to the target category via fine-tuning.
4) Compose convex point clouds for object point clouds.
5) Reconstruct the mesh surface for part-level point clouds via a pre-trained differential poison solver model (an SAP model)~\cite{peng2021shape} which is further fine-tuned on the target category.
6) Compose parts together for the final articulated mesh with a physics-aware correction scheme. 




\begin{figure*}[ht]
    \centering
      \includegraphics[width=0.7\textwidth]{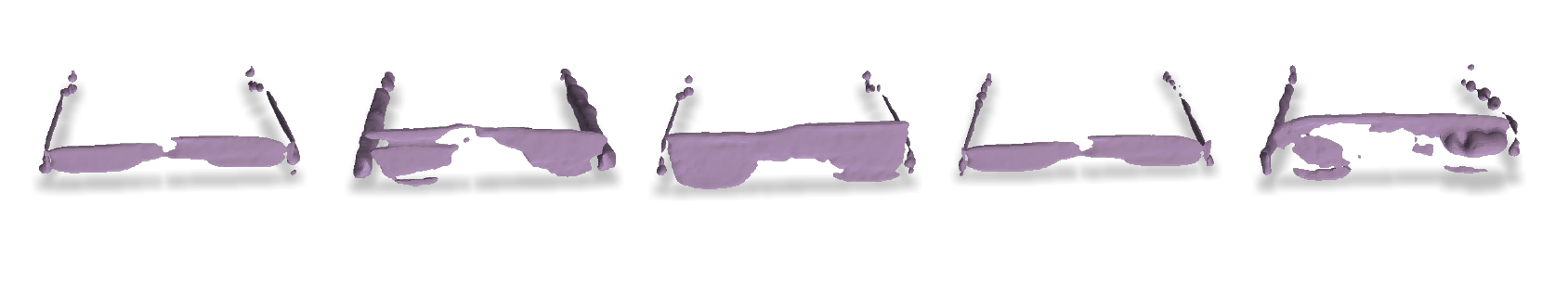}
    \caption{ \footnotesize
        \textbf{Examples of articulated meshes generated by the point cloud diffusion-based strategy} (Category: Eyeglasses). 
        } 
    \label{fig_surface_reconstruction}
\end{figure*}

\vpara{Existing problems.} 
Despite the flexible generation ability and high diversity of the sampled shapes, this strategy suffers from the poor articulated mesh quality when we further add a physics-aware correction scheme on top of reconstructed part surfaces, as shown in Figure~\ref{fig_surface_reconstruction}. 

\vpara{Discussions.} 
Generating physically-realistic articulated meshes are usually challenged by the few-shot difficulty, mesh quality, and the physically-realistic expectations. We explore a mesh deformation-based physics-aware generation strategy in this work by transferring deformation patterns from large categories and further with a physics-aware correction scheme that can improve the physical validity of generated samples while at the same time preserving transferred knowledge. 

As for two key designs in our work, 
transferring cross-category shared shape patterns at the convex level is a relatively general idea that can also be adapted to other generation techniques for shape space enrichment. However, further imposing physical validity on top of the generated samples is not a trivial thing. Our physics-aware correction works well for the deformation-based generation. But what is the most ideal strategy that can be combined with other generation techniques naturally worth further exploring. 


\section{Experimental Settings} \label{sec_supp_exp_settings}

\begin{figure*}[t]
    \centering
      \includegraphics[width=0.8\textwidth]{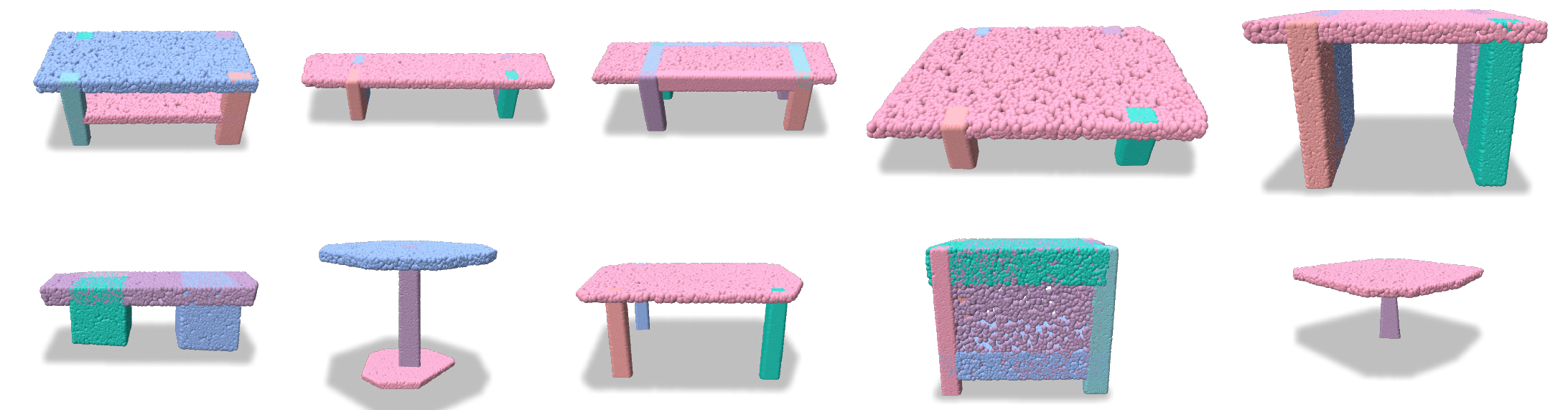}
    \caption{ \footnotesize
        \textbf{Examples for decomposed convexes} (converted to point clouds) of instances for the Table category. 
        Different colors represent different convexes. 
        } 
    \label{fig_table_cvx_decomposition}
\end{figure*}


\subsection{Datasets} \label{sec_supp_datasets}

\vpara{Rigid mesh datasets for pre-training.}
For pre-training data from rigid object categories, we select \nnpretrain instances from ShapeNet~\cite{chang2015shapenet} dataset, including Table, Chair, Lamp, and Airplane. 
We list the number of instances in each category in Table~\ref{tb_exp_nn_insts_category_rigid}. 

\vpara{Articulated mesh datasets for evaluation.}
For articulated object datasets, we select six articulated object categories from~\cite{xiang2020sapien} and four rigid object categories from ShapeNet~\cite{chang2015shapenet}. 
We list the number of its instances in Table~\ref{tb_exp_nn_insts_category}.

\begin{table*}[t]
    \centering
    \caption{\footnotesize 
    \textbf{Number of instances in each articulated object category. }
    } 
    \begin{tabular}{@{\;}c@{\;}|c|c|c|c|c|c@{\;}}
    \midrule
        \hline
        \specialrule{0em}{1pt}{0pt} 
        Method & \makecell[c]{Storage\\ Furniture} & Scissors &  Eyeglasses  & Oven & Lamp & TrashCan \\ 
        \cline{1-7} 
        \specialrule{0em}{1pt}{0pt}
        
        \#Instances & 31 & 46 & 65 & 10 & 13 & 43 
        \\ \cline{1-7} 
        \specialrule{0em}{1pt}{0pt}

    \end{tabular}

    \label{tb_exp_nn_insts_category}
\end{table*} 

\begin{table*}[htbp]
    \centering
    \caption{\footnotesize 
    \textbf{The number of instances in each rigid object category. }
    } 
    \begin{tabular}{@{\;}c@{\;}|c|c|c|c@{\;}}
    \midrule
        \hline
        \specialrule{0em}{1pt}{0pt} 
        Method & Table & Chair &  Lamp  & Airplane \\ 
        \cline{1-5} 
        \specialrule{0em}{1pt}{0pt}
        
        \#Instances & 3000 & 2447 & 1500 & 3000
        \\ \cline{1-5} 
        \specialrule{0em}{1pt}{0pt}

    \end{tabular}

    \label{tb_exp_nn_insts_category_rigid}
\end{table*}


\subsection{Baselines.} \label{sec_supp_baselines_imple}
We compare our method with two typical mesh generative strategies, namely PolyGen~\cite{nash2020polygen} that falls into the genre of direct surface generation strategies, and DeepMetaHandles~\cite{liu2021deepmetahandles} that leverages mesh deformation for generation. 
To further adapt them to our articulated mesh generation scenario, we make further modifications to their original algorithms. 
We briefly summarize our implementations of such two methods as follows. 

For PolyGen~\cite{nash2020polygen}, we download the official TensorFlow implementation. 
We then design a part-by-part generation strategy to leverage it for articulated mesh generation.
We define a part order $(\mathcal{P}_1, ..., \mathcal{P}_k)$, where $k$ is the number of parts, and generate joints of each part via generating two joint points (if any, otherwise generating no points, equivalent to generating an empty joint point set), \emph{i.e.,} $\mathcal{J}_i (1\le i\le k)$. 
Therefore, we generate mesh vertices, mesh surfaces, and joint points via the following order: $[ (\mathcal{ V}_1, \mathcal{F}_1, \mathcal{J}_1), ..., (\mathcal{V}_k, \mathcal{F}_k, \mathcal{J}_k)]$. 

For DeepMetaHandles~\cite{liu2021deepmetahandles}, we use the official implementation. 
To leverage it for articulated mesh generation, we train a deformation-based generative model for each part individually. 
Then, an object is generated by generating its part. 
Directly using the above strategy yields the default version of our compared DeepMetaHandles (``DeepMetaHandles''). 
We could further add a physics-aware deformation correction scheme on top of it, leading to the improved version (denoted as ``DeepMetaHandles w/ Phy.''). 


\subsection{Metrics.} \label{sec_supp_metrics}
For generative model-related metrics, we follow the computing processes adopted in~\cite{luo2021diffusion} and evaluate corresponding values on 4096 points sampled from mesh surfaces. 
Average Penetration Depth evaluates the average per-vertex penetration depth further averaged over all articulation states. 
Its computing process is the same as that of $\mathcal{L}_{phy}$ (see Algorithms~\ref{algo_phy_sim_single_part_single_sim}~\ref{algo_phy_sim} for details). 

\subsection{Additional Implementation Details} \label{sec_supp_addtional_imple_details}

\vpara{Convex decomposition.} 
We use BSP-Net~\cite{chen2020bsp} to provide intra-category consistent shape co-segmentation for each category. It's worth further mentioning that the number of convexes that we set for BSP-Net performs only as the upper bound of the number it uses for decomposition. 
For rigid categories, we set the number of convexes to 256. 
For articulated objects, we list their settings in Table~\ref{tb_exp_convex_nn}. 
For each part, we first try the 128 convexes setting and double the number of convexes by 2 if BSP-Net's training loss cannot converge. 

For a visual understanding of the convex decomposition, we draw some examples of decomposed convexes of instances from the Table category in Figure~\ref{fig_table_cvx_decomposition}. 

\begin{table*}[htbp]
    \centering
    \caption{\footnotesize 
    \textbf{The number of convexes used for convex decomposition} for each kind of part of each articulated object category. 
    } 
    \begin{tabular}{@{\;}c@{\;}|c|c|c|c|c|c@{\;}}
    \midrule
        \hline
        \specialrule{0em}{1pt}{0pt} 
        Method & \makecell[c]{Storage\\ Furniture} & Scissors &  Eyeglasses  & Oven & Lamp & TrashCan \\ 
        \cline{1-7} 
        \specialrule{0em}{1pt}{0pt}
        
        Link 0 & 128 & 128 & 128 & 128 & 128 & 128 
        \\ \cline{1-7} 
        \specialrule{0em}{1pt}{0pt}

        Link 1 & 512 & 128 & 128 & 512 & 128 & 512 
        \\ \cline{1-7} 
        \specialrule{0em}{1pt}{0pt}

        Link 2 & 512 & - & 128 & - & 128 & - 
        \\ \cline{1-7} 
        \specialrule{0em}{1pt}{0pt}

         Link 3 & - & - & - & - & 128 & - 
        \\ \cline{1-7} 
        \specialrule{0em}{1pt}{0pt}

    \end{tabular}
    \vspace{-16pt}
    \label{tb_exp_convex_nn}
\end{table*}

\vpara{Convex-level generative model.}
For the convex-level generative model, we use a sphere mesh containing 42 vertices and 80 faces~\cite{yifan2020neural} as the template of cages. 
The number of deformation bases is set to 16. 

We use a neural network $\psi_\theta(\cdot)$ to parameterize deformation bases. 
It takes a convex $c$ as input and predicts its deformation bases $B_c = \psi_\theta(c)$. 
It first extracts the convex feature for $c$ via a PointNet encoder (applied on 4096 points sampled from $c$'s surface). 
Then we feed the convex feature and the cage $t_c$ of the convex $c$ to a MultiFold network, same as the network used in~\cite{yifan2020neural} for per-point features with the bottleneck size set to 512 and the number of folds set to 3. 
We then use an MLP for basis prediction with weight size in each layer set to (128, 128), (128, 128), and (128, 48) with LeakyReLU layer using the default $\alpha$ value between every two fully-connected layers. 


\vpara{Training protocols.}
In the pre-training stage, total losses for optimizing the convex-level deformation-based generative model is composed of the convex deformation loss $\mathcal{L}_C$ and two penalty terms, \emph{i.e.,} $\mathcal{L}_{sp}$ and $\mathcal{L}_{orth}$. 
Specifically, the total loss is $\mathcal{L} = \mathcal{L}_C + \lambda_{sp}\cdot \mathcal{L}_{sp} + \lambda_{orth} \cdot \mathcal{L}_{orth}$, where $\lambda_{sp}$ and $\lambda_{orth}$ are both set to $10^{-4}$ in our implementation. 

In the fine-tuning stage, the object-level physical penalty $\mathcal{L}_{phy}$ (calculated on the final shape optimized via $\mathcal{L}_{proj}$) is further added to encourage the network to produce physically-realistic deformations. 
Therefore, the total loss becomes $\mathcal{L} = \mathcal{L}_C + \lambda_{phy}\cdot \mathcal{L}_{phy} + \lambda_{sp}\cdot \mathcal{L}_{sp} + \lambda_{orth} \cdot \mathcal{L}_{orth}$. 
We empirically set $\lambda_{phy}$ to 1.0. 

We use the Adam optimizer with the momentum set to  (0.9, 0.999) for optimization in both the pre-training and the fine-tuning stages. 
The initial learning rate is set to $10^{-4}$. 
And it decayed by 0.5 after every 100 epochs.



\vpara{Evaluation protocols.}
We adopt the few-shot generation evaluation strategy. 
The default number of shots is set to 5. 
We use the same test set to compare different methods. 
Specifically, for each articulated mesh category, we randomly select 5 instances for training while using the remaining instances for test. 
At the test time, such 5 instances serve as reference examples to generate new samples. 
For each reference example, we generate 40 samples from it. 
It is realized by passing the example mesh into the hierarchical deformation generative model and randomly sampling 40 global deformation coefficients $\mathbf{z}$ from its deformation coefficient distribution model.
Each global deformation coefficient $\mathbf{z}$ together with the synchronized deformation bases $\{ S_c B_c \}$ are used to deform the example to a new shape. 

As for the 2-shot, 4-shot, and 8-shot settings present in the supplementary material, we use the same strategy to split instances in the articulated mesh category into a few-shot training set and a test set. 

\end{document}